\documentclass{article}
\usepackage{arxiv}

\usepackage{mathtools} 

\usepackage{booktabs}
\usepackage{array,multirow}
\usepackage{rotating}
\usepackage{makecell}
\usepackage{rotating}
\usepackage{array}
\usepackage{float}
\usepackage{physics}
\usepackage{algorithm}
\usepackage{algpseudocode}
\usepackage{enumitem}
\usepackage[font={small,it}]{caption}
\usepackage{hyperref}
\usepackage{microtype}

\usepackage{amssymb}
\usepackage{amsmath}
\usepackage{mathdots}
\usepackage{mathtools}
\usepackage{multicol}

\usepackage{tensor}

\usepackage{urwchancal}
\DeclareFontFamily{OT1}{pzc}{}
\DeclareFontShape{OT1}{pzc}{m}{it}{<-> s * [1.10] pzcmi7t}{}
\DeclareMathAlphabet{\mathpzc}{OT1}{pzc}{m}{it}

\usepackage{graphicx}
\usepackage{ifpdf}

\ifpdf
\usepackage{epstopdf}
\PrependGraphicsExtensions{.svg}
\fi

\usepackage{booktabs}
\usepackage{array,multirow}
\usepackage{rotating}
\usepackage{makecell}
\usepackage{rotating}
\usepackage{array}
\usepackage{float}
\usepackage{physics}
\usepackage{algorithm}
\usepackage{algpseudocode}
\usepackage{bbding}
\usepackage{amsmath}
\usepackage{bm}

\usepackage{wrapfig}

\newcommand{\eg}{\textit{e}.\textit{g}., }

\providecommand{\idx}[5]{\tensor*[_{#3}^{#2}]{#1}{_{#4}^{#5}}}
\usepackage{adjustbox}
\usepackage{multirow}
\usepackage[dvipsnames]{xcolor}
\usepackage[normalem]{ulem}

\newcommand{\publicationdetails}
{Wang, Z., Pan, L., Ng, Y., Zhuang, Z., Mahony R. (2021), ``Stereo Hybrid Event-Frame (SHEF) Cameras for 3D Perception", \textit{published in IEEE International Conference on Intelligent Robots and Systems (IROS), 2021, pp. 9758-9764, DOI: \href{https://ieeexplore.ieee.org/document/9636312}{10.1109/IROS51168.2021.9636312}}. Liyuan Pan and Yonhon Ng contribute equally. \\
\copyrightNoticeIEEE{2021}	\\
	}
\newcommand{\publicationversion}
{Author Accepted Version}

\title{\LARGE \bf
	Stereo Hybrid Event-Frame (SHEF) Cameras for 3D Perception}

\author{
	\href{https://orcid.org/0000-0003-0815-1287}{\includegraphics[scale=0.06]{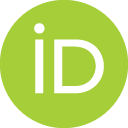}\hspace{1mm}
		Ziwei Wang}
	\\
	Systems Theory and Robotics Group \\
	Australian National University \\
	ACT, 2601, Australia \\
	\texttt{ziwei.wang1@anu.edu.au} \\
	\\
	\And
	\href{https://orcid.org/0000-0002-9025-173X}{\includegraphics[scale=0.06]{orcid.png}\hspace{1mm}
		Liyuan Pan$^{*}$}
	\\
	Agriculture and Food, CSIRO \\
	Australian National University \\
	ACT, 2601, Australia \\
	\texttt{liyuan.pan@anu.edu.au} \\
	\And
	\href{https://orcid.org/0000-0002-7764-298X}{\includegraphics[scale=0.06]{orcid.png}\hspace{1mm}
		Yonhon Ng$^{*}$}
	\\
	Systems Theory and Robotics Group \\
	Australian National University \\
	ACT, 2601, Australia \\
	\texttt{yonhon.ng@anu.edu.au} \\
	\\
	\And
	\href{https://orcid.org/0000-0002-6632-3342}{\includegraphics[scale=0.06]{orcid.png}\hspace{1mm}
		Zheyu Zhuang}
	\\
	Systems Theory and Robotics Group \\
	Australian National University \\
	ACT, 2601, Australia \\
	\texttt{zheyu.zhuang@anu.edu.au} \\
	\And	\href{https://orcid.org/0000-0002-7803-2868}{\includegraphics[scale=0.06]{orcid.png}\hspace{1mm}
		Robert Mahony}
	\\
	Systems Theory and Robotics Group \\
	Australian National University \\
	ACT, 2601, Australia \\
	\texttt{robert.mahony@anu.edu.au} \\
}

\begin{document}

\maketitle
\thispagestyle{empty}
\pagestyle{empty}


\begin{abstract}
	Stereo camera systems play an important role in robotics applications to perceive the 3D world.
	However, conventional cameras have drawbacks such as low dynamic range, motion blur and latency due to the underlying frame-based mechanism.
	Event cameras address these limitations as they report the brightness changes of each pixel independently with a fine temporal resolution, but they are unable to acquire absolute intensity information directly.
	Although integrated hybrid event-frame sensors (\eg DAVIS) are available, the quality of data is compromised by coupling at the pixel level in the circuit fabrication of such cameras. 
	This paper proposes a stereo hybrid event-frame (SHEF) camera system that offers a sensor modality with separate high-quality pure event and pure frame cameras, overcoming the limitations of each separate sensor and allowing for stereo depth estimation.
	We provide a SHEF dataset targeted at evaluating disparity estimation algorithms and introduce a stereo disparity estimation algorithm that uses edge information extracted from the event stream correlated with the edge detected in the frame data.
	Our disparity estimation outperforms the state-of-the-art stereo matching algorithm on the SHEF dataset.
\end{abstract}

\centerline{
	\noindent \textbf{Code, Datasets and Video:}
}
\centerline{
	\noindent \href{https://github.com/ziweiWWANG/SHEF.git}{{\color{pink}\texttt{https://github.com/ziweiWWANG/SHEF.git}}}
}

\section{INTRODUCTION}
Event cameras based on bio-inspired retina vision sensor, such as Dynamic Vision Sensor (DVS)~\cite{Lichtsteiner08ssc} asynchronously measure per-pixel log intensity changes with a fine time resolution ($<$10$\mu$s), low latency ($<$0.5ms), high dynamic range ($>$120dB vs 60dB of conventional cameras), and low power consumption ($<$0.1W)~\cite{Lichtsteiner08ssc,brandli2014240,hu2021v2e}.
These properties make event cameras of significant interest in robotics applications where the robustness of the sensing system is crucial for challenging scenarios with fast motion and high contrast scenes.
However, event cameras do not capture the absolute intensity and static image information. 
There is considerable potential for robotic systems to exploit the complementary characteristics of both event and frame-based cameras, motivating the consideration of hybrid and stereo hybrid event-frame sensor systems. 
The dynamic and active pixel vision sensor (DAVIS)~\cite{brandli2014240} combines the DVS and an active pixel sensor (APS) circuit.
By sharing the same photodiode, DAVIS provides registered event data and frames at the pixel level that allows easy access to dual event-frame data. 
However, the coupling between APS and DVS components generates biased event noise triggered by the shutter of each frame~\cite{brandli2014240}.
This hybrid configuration problem is significant, and limits the quality of both event and frame data.
Currently, the highest quality of the hybrid event-frame camera is DAVIS346 with a limited resolution of 346 $\times$ 260 in both event data and frames. 
In contrast, pure event cameras have achieved 1280 $\times$ 960 resolution in Samsung Gen4 and are developing more quickly than the hybrid event-frame sensors.
The stereo hybrid event-frame sensor modality offers significant performance advantages as well as allowing simple integration into existing sensor suites, such as the multi-camera systems commonly found on modern mobile phones and robotic systems. 
In addition, adding event sensors to existing robotic systems can compensate for failure of an existing conventional frame camera in low-light or high-velocity scenarios. 
However, in order to exploit event data with existing systems, it is necessary to estimate disparity between the separate event sensor and the existing frame sensor. 
Such a disparity, and consequently depth, estimate will also provide additional sensing capability for future robotic systems.

\begin{figure}[t!]
	\centering
	\resizebox{0.65\columnwidth}{!}{
		\begin{tabular}{c c}
			\\	\includegraphics[width=0.5\linewidth]{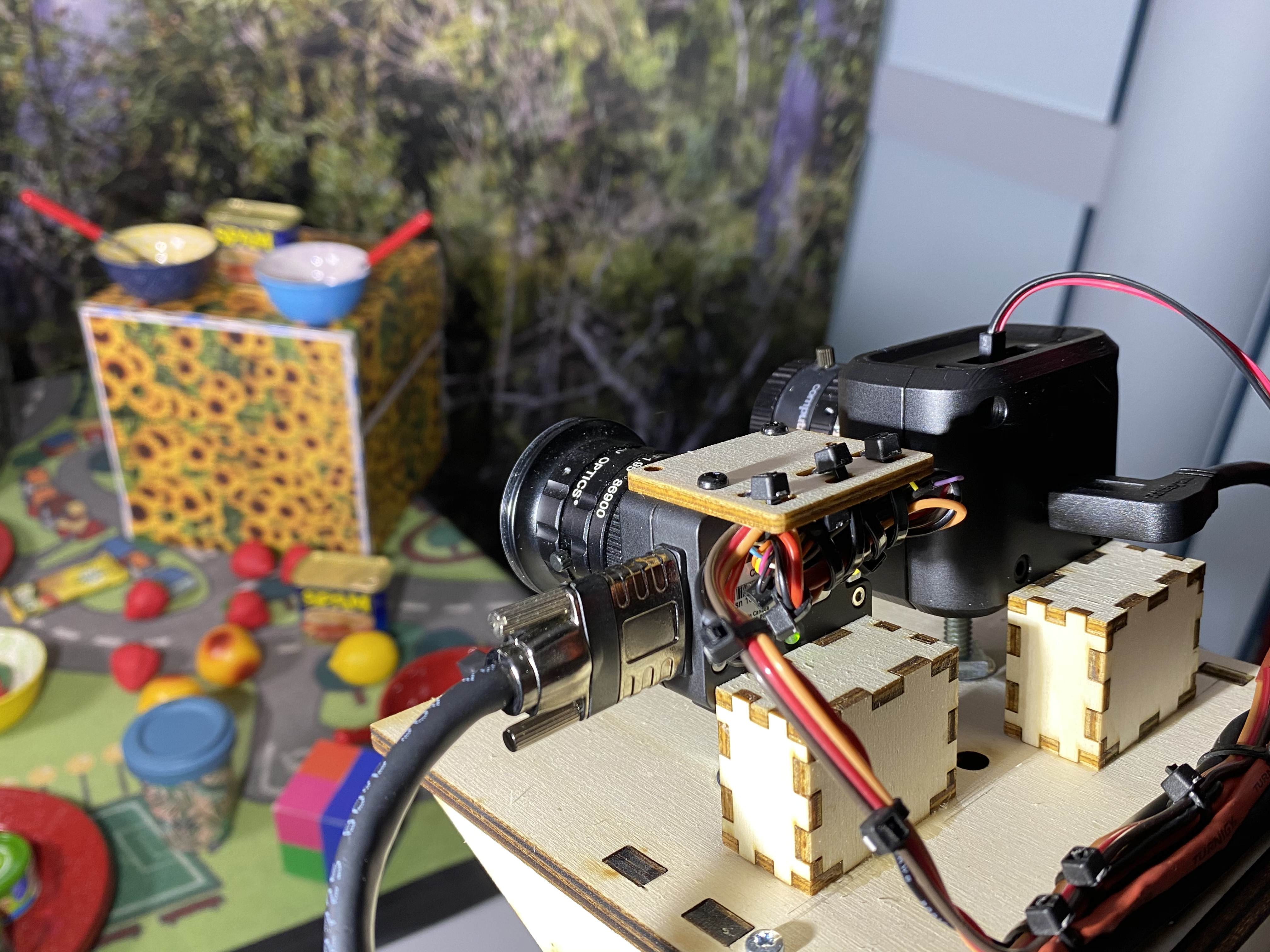} &
			\includegraphics[width=0.5\linewidth]{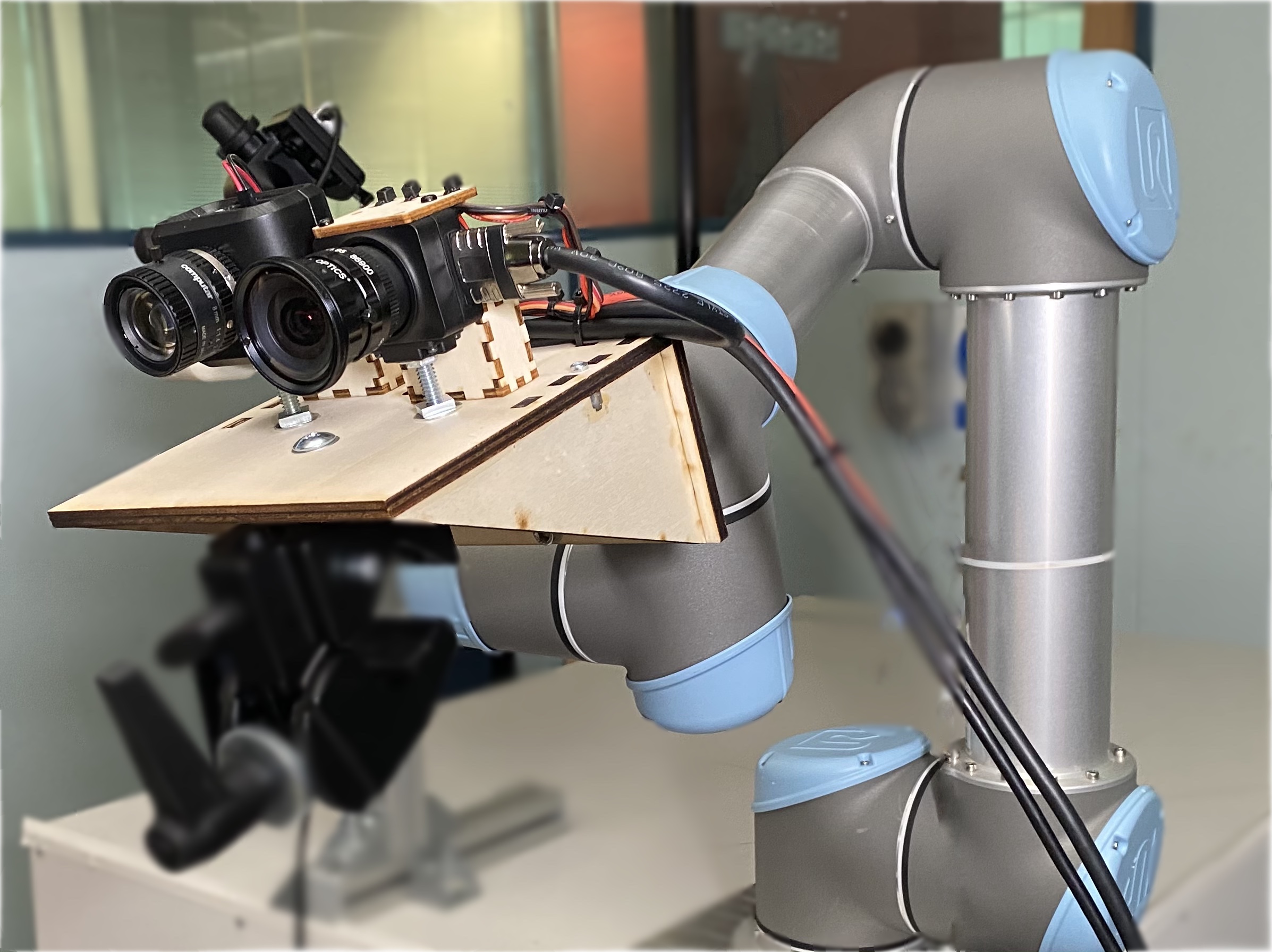} \\
			(a) SHEF system & (b) SHEF system mounted on UR5 \\
			\includegraphics[width=0.5\linewidth]{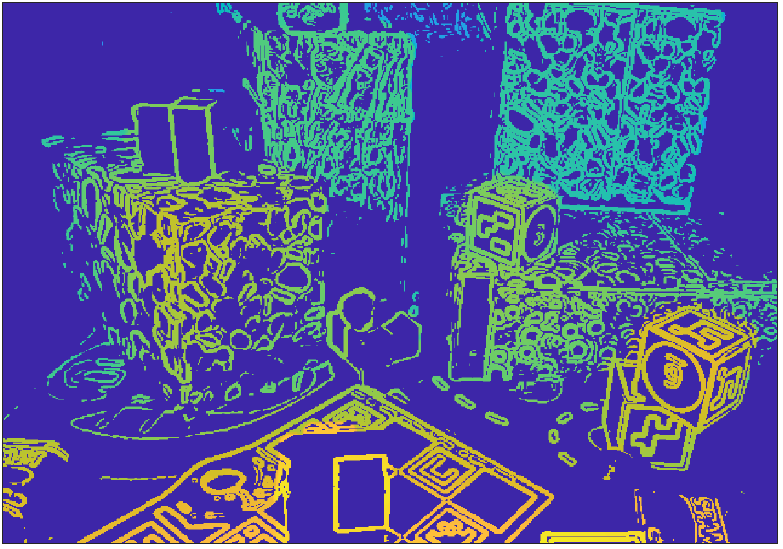} &
			\includegraphics[width=0.5\linewidth]{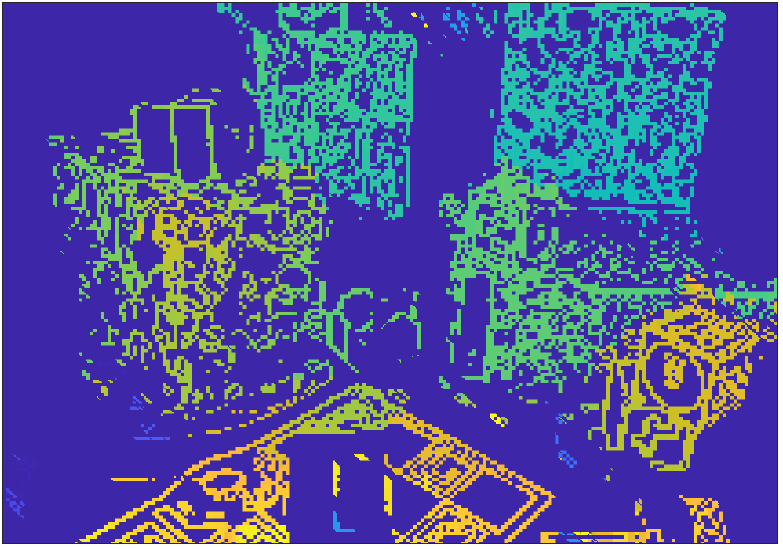} \\
			(c) Ground truth disparity & (d) Our disparity \\
		\end{tabular}
	}
	\caption{\label{fig:hardware} Concept of our Our stereo hybrid event-frame (SHEF) camera system.
		(a) Our SHEF camera system collects a dataset for the stereo hybrid event-frame depth estimation task.
		(b) The stereo camera system is mounted on a UR5 manipulator, providing controlled camera motion and accurate poses.
		(c) An example of the ground truth disparity of the image edge map.
		(d) Our estimated disparity map $\mathbf{D}_p$ 
		corresponding to the provided ground truth disparity. It is computed by our cross-correlation method using edge information from event data and frame data.
	}
\end{figure}

To the authors' best knowledge, this paper is the first work that tackles the problem of fusing event and frame data (separate sensing modalities) to compute stereo disparity.
In this paper, we introduce the first dataset targeted specifically at stereo hybrid event-frame data for near field scenes and a baseline algorithm for computing disparity.
Our stereo hybrid event-frame (SHEF) camera system consists of a separate event camera and a frame-based camera alongside each other.
We mount the system rigidly on a UR5 robot manipulator to get the controlled motion and accurate camera poses. 
Then we build the ground truth 3D environment reconstruction using a sequence of RGB images with known camera positions from the UR5.
Our baseline (edge-based) disparity estimation algorithm computes cross-correlation between event data and frames with a coarse-to-fine framework.
As an indication of the potential of our system, we also train a Disparity Completion Net (DCNet) to interpolate and optimize the sparse disparity map to generate a dense map, taking advantage of the absolute intensity information from the frame data. 
Our dataset and code will be made available online for future studies and comparisons.

In summary, the contributions of the paper are:
\begin{itemize}
	\item Formulate the multi-modality stereo hybrid event-frame depth estimation problem and introduce a baseline disparity estimation algorithm. 
	\item
	Introduce the first open-source high-quality stereo hybrid event-frame dataset targeted specifically at depth estimate for near field scenes,
	including event data, high frame rate RGB image sequences, accurate camera poses, point clouds of 3D environment reconstructions and ground truth depth maps.
\end{itemize}

\section{RELATED WORK}
Since the introduction of integrated hybrid event-frame cameras, in particular the DAVIS cameras, many datasets that incorporate both event and frame data are developed. 
These datasets target a wide range of applications, such as
optical flow~\cite{pan2020single}, 
tracking~\cite{tedaldi2016feature, liu2016combined, Gehrig19ijcv}, video reconstruction~\cite{Brandli14iscas, Scheerlinck18accv,Wang2019, Pan20pami,wang2021asynchronous} and
deblurring~\cite{jiang2020learning, pan2019bringing}.
However, the hybrid DAVIS cameras 
generate biased event noise triggered by the shutter of each frame when operating in hybrid mode.
Some researchers provide datasets with separate event-frame sensors. 
Wang \textit{et~al.}~\cite{wang2020joint} propose a hybrid system with a DAVIS 240b event camera (resolution of 180 $\times$ 190, operating in pure event mode) and a FLIR RGB frame-based camera. Events and frames are registered through a beam-splitter.
The system overcomes the inherent noise of DAVIS and acquires high-resolution frames from a separate 
commercial frame sensor.
Despite having two separate sensors, the
system is effectively a monocular set-up and it only provides relatively low-resolution event data from the DAVIS 240b event camera. 
Along with other on-board sensors, stereo DAVIS system proposed by Zhu \textit{et al.}~\cite{zhu2018multivehicle} provide stereo event data, stereo frames, camera poses, reference depth images from embedded GPS, VI sensor and Lidar.
Because the system provides depth ground truth, many recent stereo event-based 3D perception works are evaluated on the released dataset of this system, \eg \cite{Zhou18eccv,tulyakov2019learning,hadviger2020stereo}.
However, this dataset is inherently affected by bias and shutter noise due to the use of DAVIS's hybrid event-frame mode~\cite{brandli2014240}.
Furthermore, in the widely-used indoor flying sequences for the 3D perception tasks~\cite{zhu2018multivehicle}, the event data is sparse and heavily biased (the ratio of positive to negative events is around 2.5-5)~\cite{zhu2018multivehicle}.
Recent datasets that use separate event and frame cameras \cite{Gehrig21ral,tulyakov2021time,Zou_2021_CVPR,wang2021asynchronous} have been shown to lead to higher quality depth estimation and image reconstruction.

A number of authors have considered stereo event-event and monocular ``multi-view'' event depth estimation as well as using event data to augment frame based stereo or multi-view depth estimation. 
Event-driven stereo matching methods were proposed in \cite{zou2017robust} and~\cite{camunas2017event}.
The disparity uniqueness and depth continuity in stereo event depth estimation has also been considered by~\cite{firouzi2016asynchronous} and \cite{xie2017event}.
Osswald \emph{et al.}~\cite{Osswald17srep} propose a spiking stereo neural network to compute disparity that can be implemented in parallel.
Zhou \emph{et al.}~\cite{Zhou18eccv} compute a semi-dense depth map from stereo events at multiple viewpoints.
Monocular multi-view event-based depth estimation algorithms are usually used to solve SLAM and visual odometry tasks~\cite{Rebecq17ral,Rebecq18ijcv,Gallego18cvpr}, computing relative camera poses and 3D reconstructions over time.
Mitrokhin \emph{et al.}~\cite{mitrokhin2019ev} present a dataset with accurate motion masks and ground truth depth that can be used in monocular event-based depth estimation.
To exploit the advantage of different sensing modalities, Hadviger \textit{et~al.}~\cite{hadviger2020stereo} compute dense disparity from frames, and predict future disparity with visual odometry, then update disparity using event-based optical flow.

\begin{table}
	\vspace{3mm}
	\centering
	\caption{\label{tab:dataset} Summary of our stereo hybrid event-frame dataset.}
	\resizebox{0.6\textwidth}{!}{ 
		\begin{tabular}{ l | l }
			\midrule
			\midrule
			Sensors & Prophesee VGA event sensor 
			\\ & FLIR Chameleon3USB3 54fps RGB frame sensor
			\\
			\midrule
			Environments & simple boxes, complex boxes and picnic
			\\
			\midrule
			Lighting conditions & night + single LED light 
			\\ & day light + two LED lights
			\\
			\midrule
			Camera motion & square + circle + three Lissajous curves
			\\
			\midrule
			End effector speed range & low speed \hspace{4.2mm} : 0.01 to 0.14 m/s \\ &
			median speed :  0.02 to 0.25 m/s \\ &
			high speed  \hspace{3.2mm} : 0.03 to 0.35m/s 
			\\
			\midrule
			Depth range & 1.5 to 4.5 meters
			\\
			\midrule
			Ground truth & camera and UR5 manipulator poses sampled at 125 Hz
			\\ & dense depth ground truth for each frame
			\\ & point clouds for each scene
			\\ & 3D reconstruction model for each scene
			\\
			\midrule
			Camera parameters & stereo camera intrinsic and extrinsic parameters \\ & hand-eye transformation parameters
			\\
			\midrule
			Size & more than 55000 RGB images sampled at 54 pfs \\ & 70 GB raw event data \\
			\midrule
			\midrule
		\end{tabular}
	}\label{table: rig dataset}
\end{table} 

\section{Stereo Hybrid Event-frame Dataset}
\begin{figure*}
	\centering
	\begin{tabular}{cccc}
		\hspace{-0.35 cm}
		\includegraphics[width=0.33\textwidth]{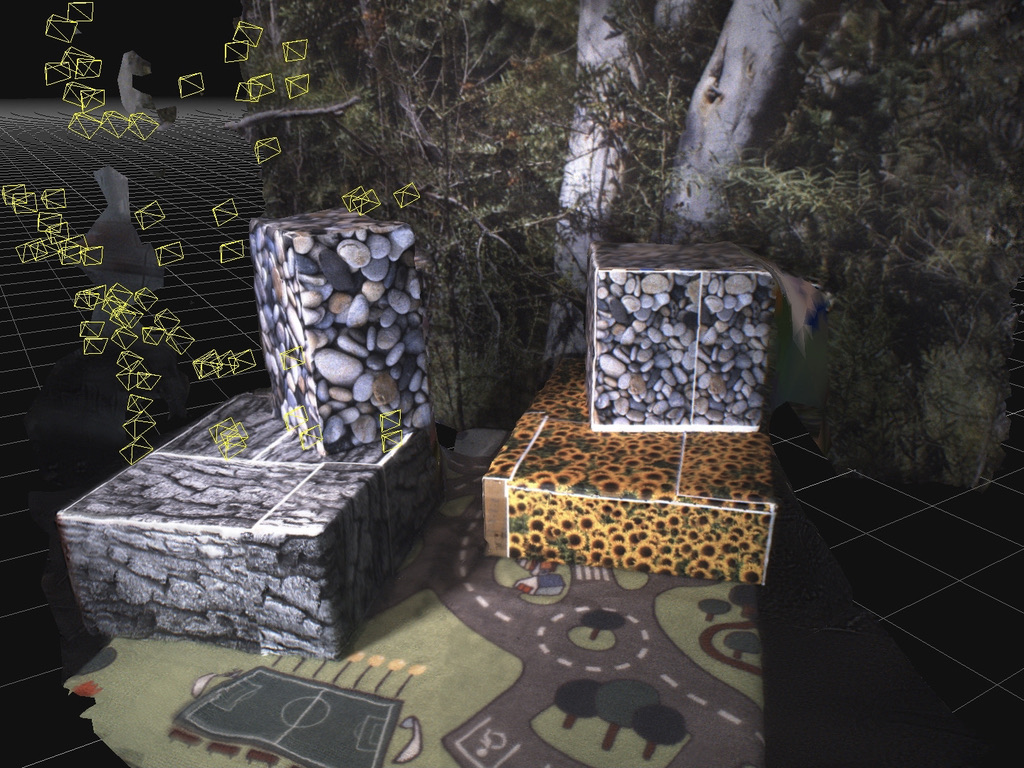} 
		\hspace{-0.35 cm}
		&  \includegraphics[width=0.33\textwidth]{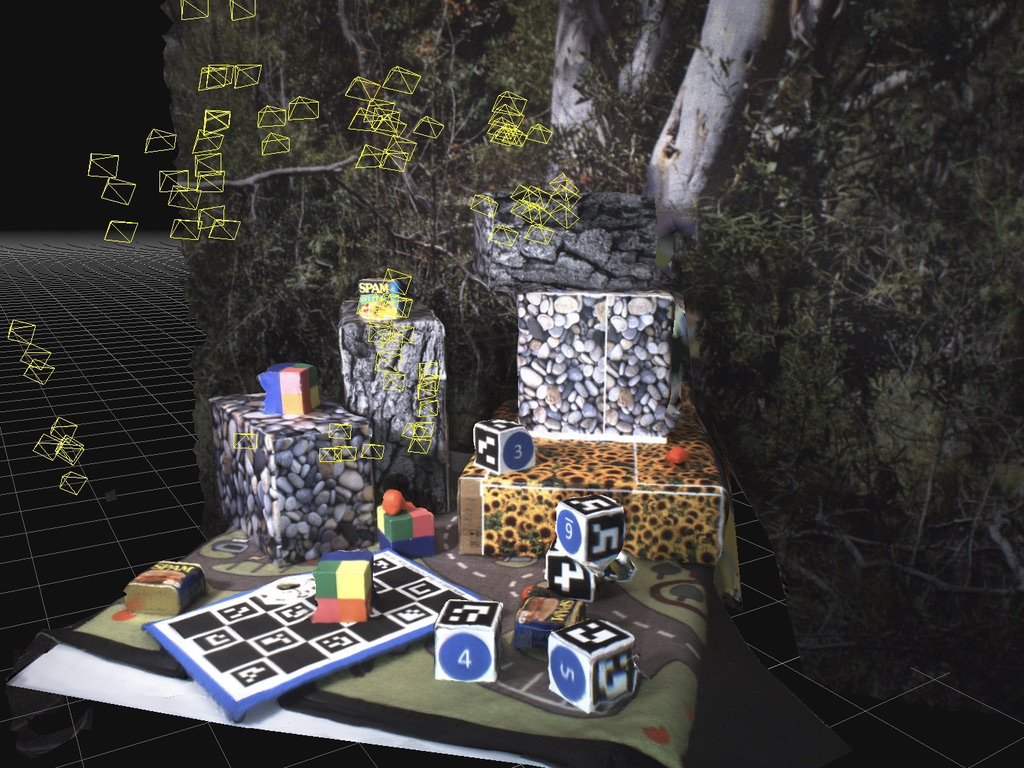} 
		\hspace{-0.35 cm}
		& \includegraphics[width=0.33\textwidth]{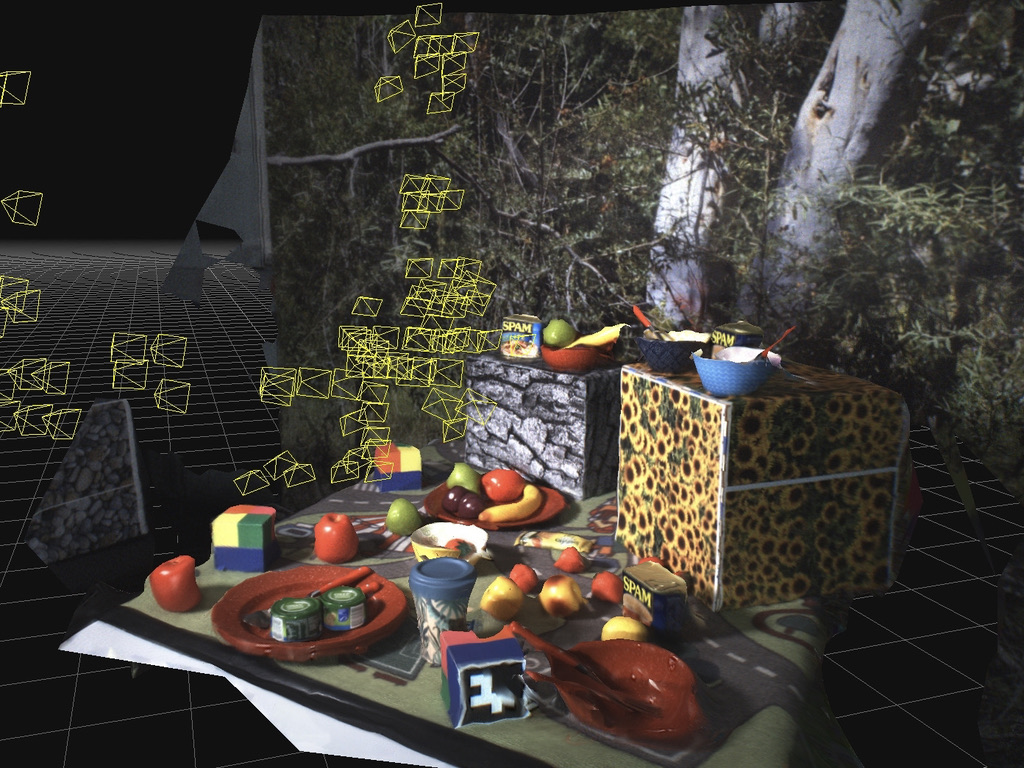} \\
		\hspace{-0.35 cm}
		(a) \texttt{Simple boxes}
		\hspace{-0.35 cm}
		& (b) \texttt{Complex boxes}
		\hspace{-0.35 cm}
		&(c) \texttt{Picnic}   \\
	\end{tabular}
	\caption{The 3D reconstruction model of our three dataset scenes.
		For each scene, we use the FLIR RGB camera in our system to take multiple pictures from various viewpoints.
		Accurate camera poses (shown in yellow) with respect to the robot base are computed by the camera to robot end-effector transformation matrix and the forward kinematics model of the UR5 manipulator.
	}
	\label{fig:gt}
\end{figure*}

In this section, we introduce the proposed SHEF system and provide details of the dataset. 
It covers the sensor parameters, sampling rate, hardware set up, three dataset scenes, camera motion control, system calibration, 3D reconstruction model and depth ground truth.
\subsection{Methodology}
\noindent{\bf{Sensors: }}
We build a stereo hybrid event-frame camera system with a Prophesee event camera (VGA, $640\times480$ pixels) and a FLIR RGB camera (Chameleon3USB3, $2048\times1536$ pixels, 54fps) mounted side-by-side on a camera rig. This rig is then mounted on the end-effector of a UR5 robot manipulator during the data collection to obtain accurate camera poses with respect to the robot base.
The baseline $\mathbf{b}$ and focal lengths $\mathbf{f}$ of the stereo set up are:
$\mathbf{b} = 65.44$ mm, $\mathbf{f} = 555$ pixels for event camera and $\mathbf{f} = 1301$ pixels for FLIR camera.

\noindent{\bf{Environment: }}
The presented dataset includes three different scenes: (a) \texttt{simple boxes} is the baseline dataset consisting of four big boxes with different sizes and depths on a carpet.
(b) \texttt{complex boxes} includes five big boxes and eight small cubes, spam cans, strawberries, a chessboard with ArUco markers at different depths.
(c) \texttt{picnic} is the most challenging dataset we provide. It includes all kinds of fruit, tableware and cutlery,  which are small, reflective and tend to cause ambiguities by object occlusion.
The boxes, background and table surface are highly textured to improve the quality of the dataset for evaluating depth and disparity reconstruction. 
Each scene in the presented dataset includes three separate data classes: events, RGB frames, and camera trajectories of both the event and the frame-based cameras. The RGB frames are sampled at 54 fps and the camera trajectories are sampled at approximately 125 Hz.

\begin{figure}
	\centering
	\begin{tabular}{cccc}
		\hspace{-0.35 cm}
		\includegraphics[width=0.21\textwidth]{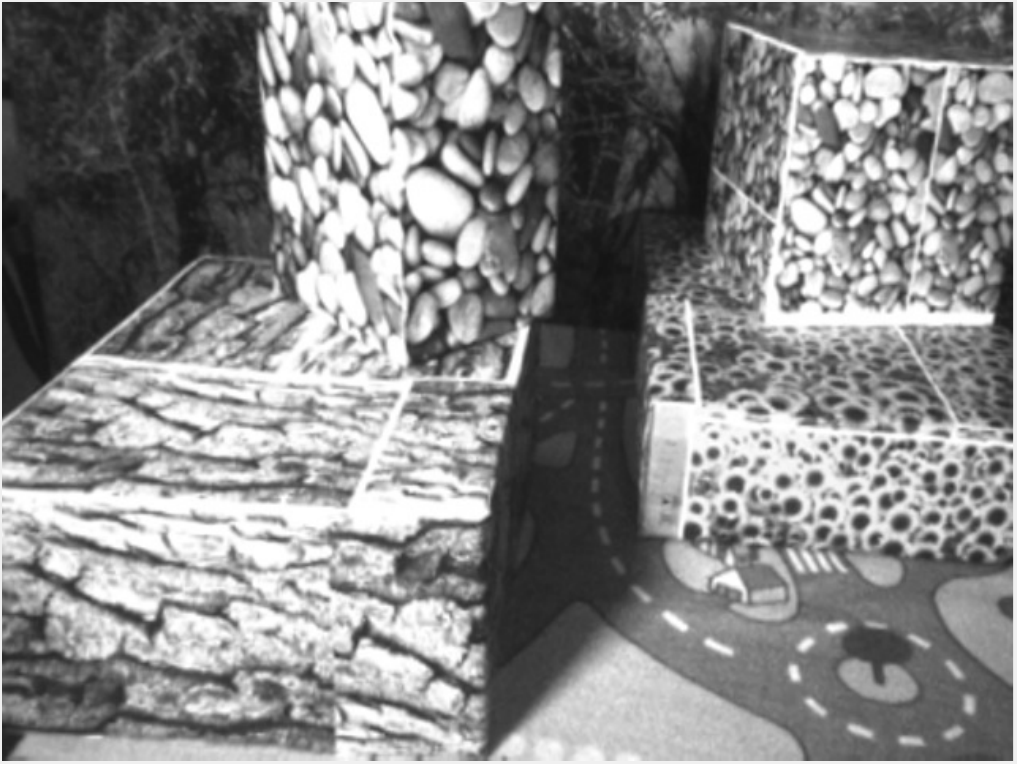}
		\hspace{-0.5 cm}
		& \includegraphics[width=0.21\textwidth]{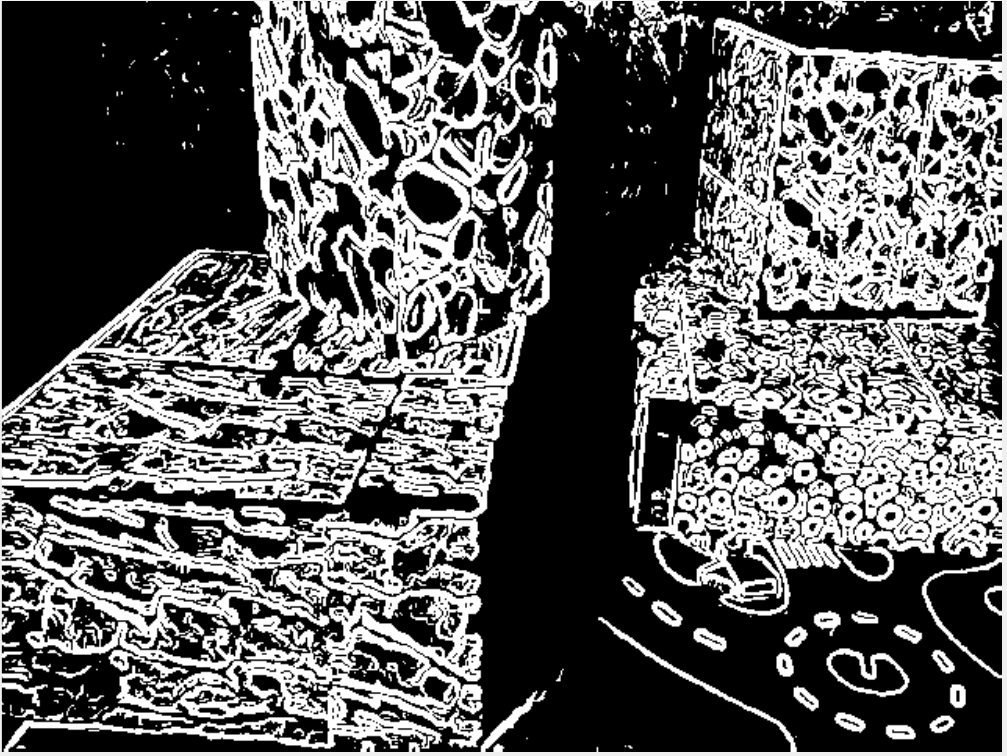}
		\hspace{-0.5 cm}
		& \includegraphics[width=0.21\textwidth]{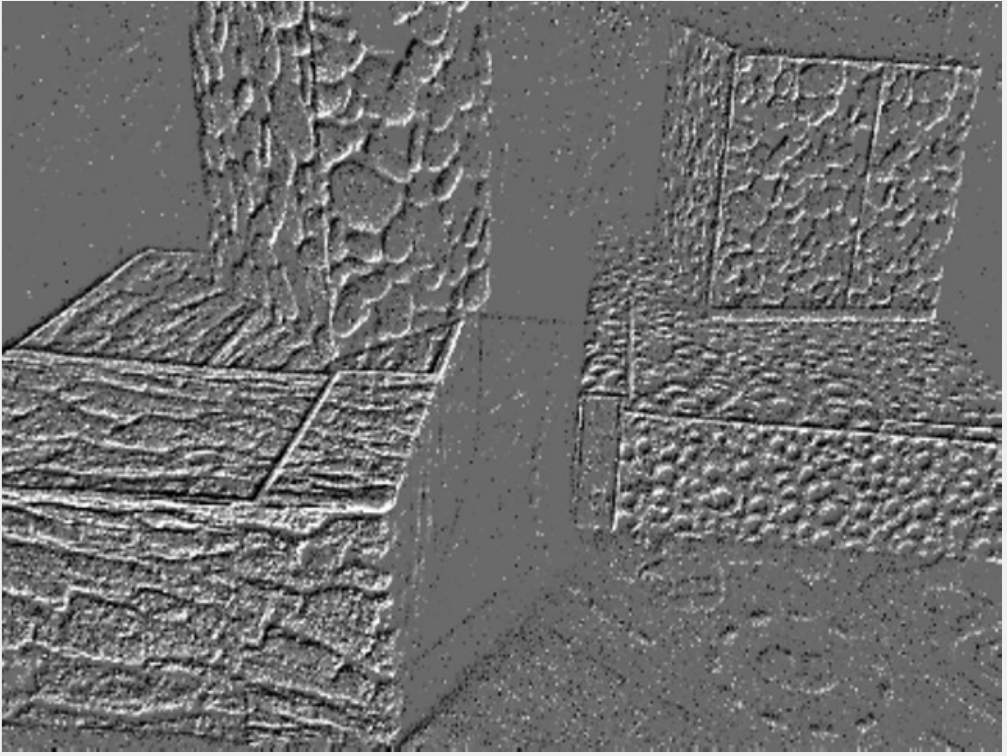}
		\hspace{-0.5 cm}
		& \includegraphics[width=0.21\textwidth]{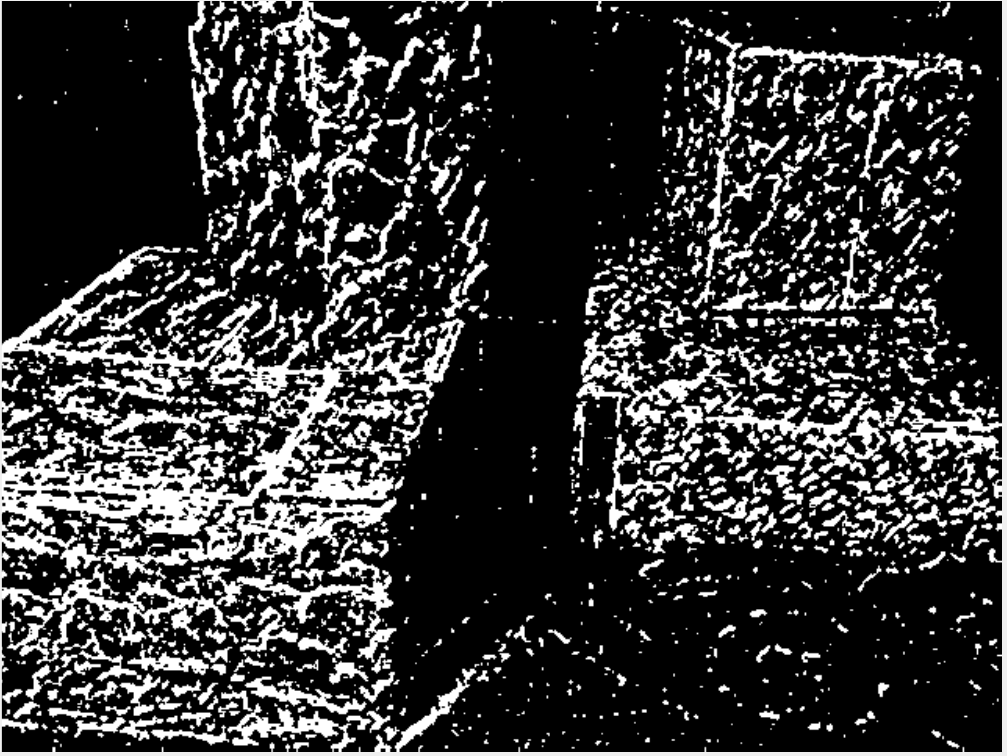}\\
		\hspace{-0.35 cm}
		\includegraphics[width=0.21\textwidth]{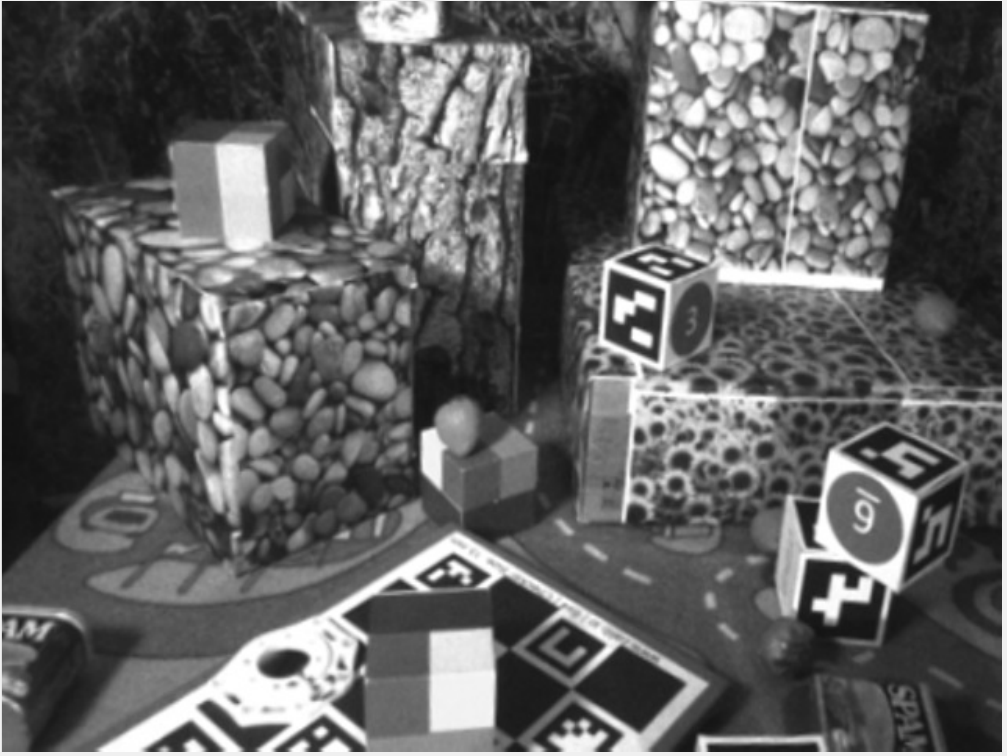}
		\hspace{-0.5 cm}
		& \includegraphics[width=0.21\textwidth]{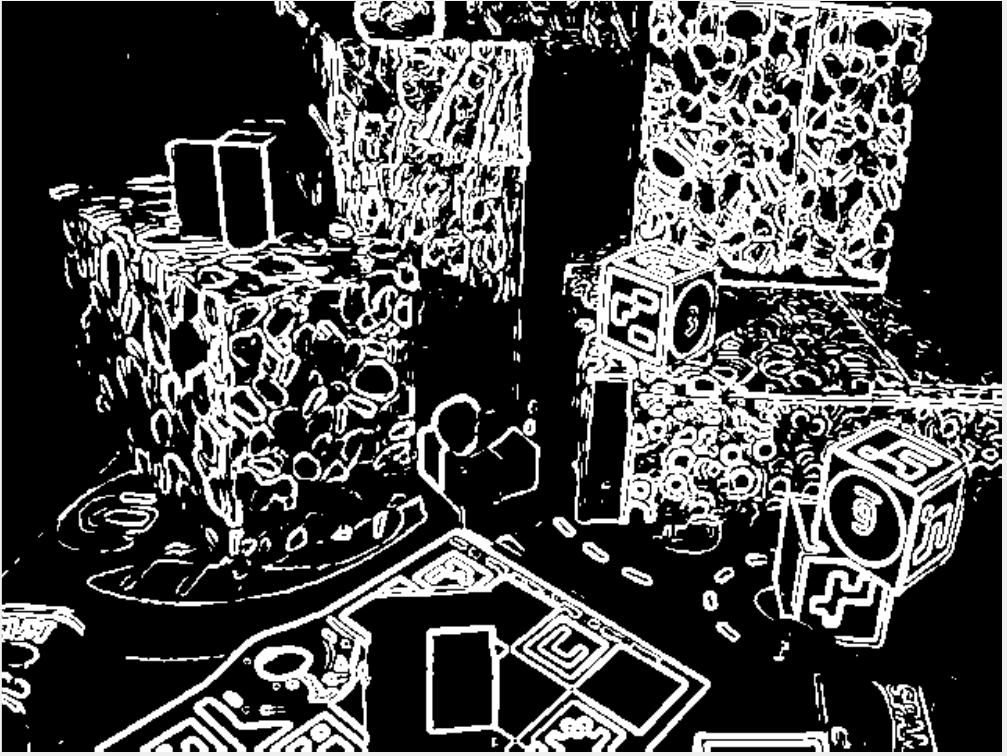}
		\hspace{-0.5 cm}
		& \includegraphics[width=0.21\textwidth]{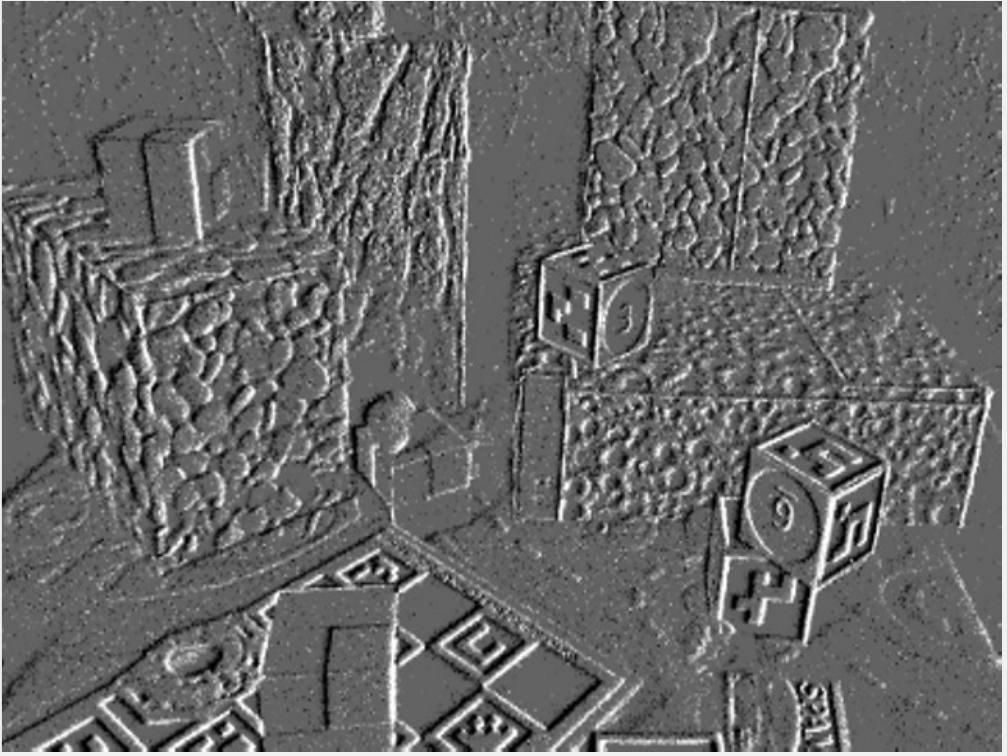}
		\hspace{-0.5 cm}
		& \includegraphics[width=0.21\textwidth]{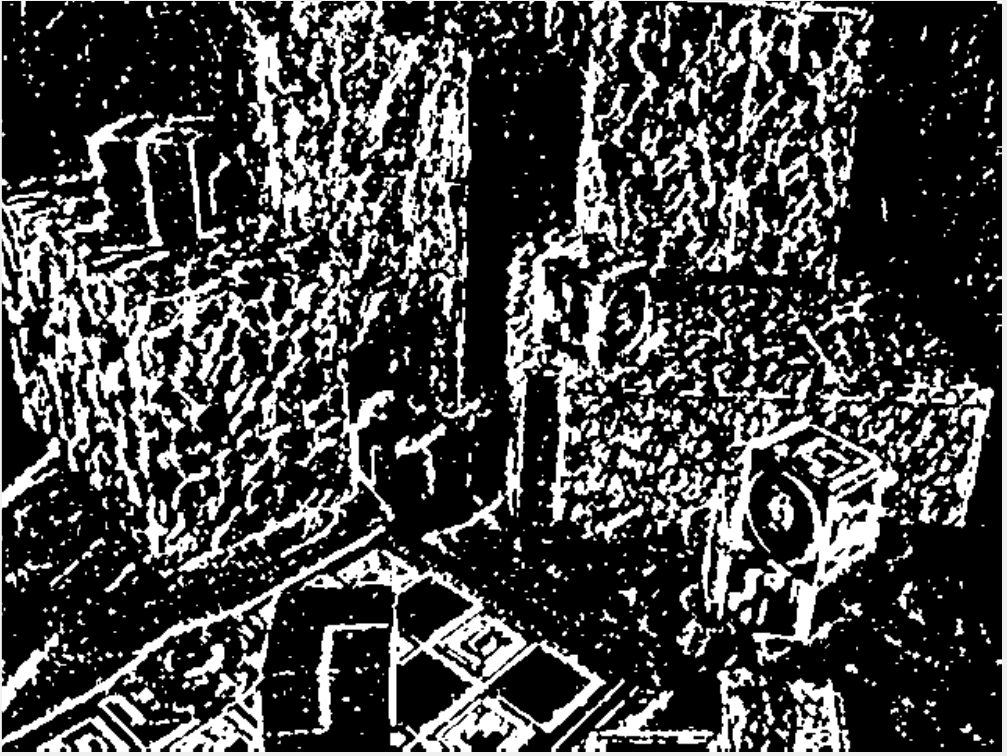} \\
		\hspace{-0.35 cm}
		\includegraphics[width=0.21\textwidth]{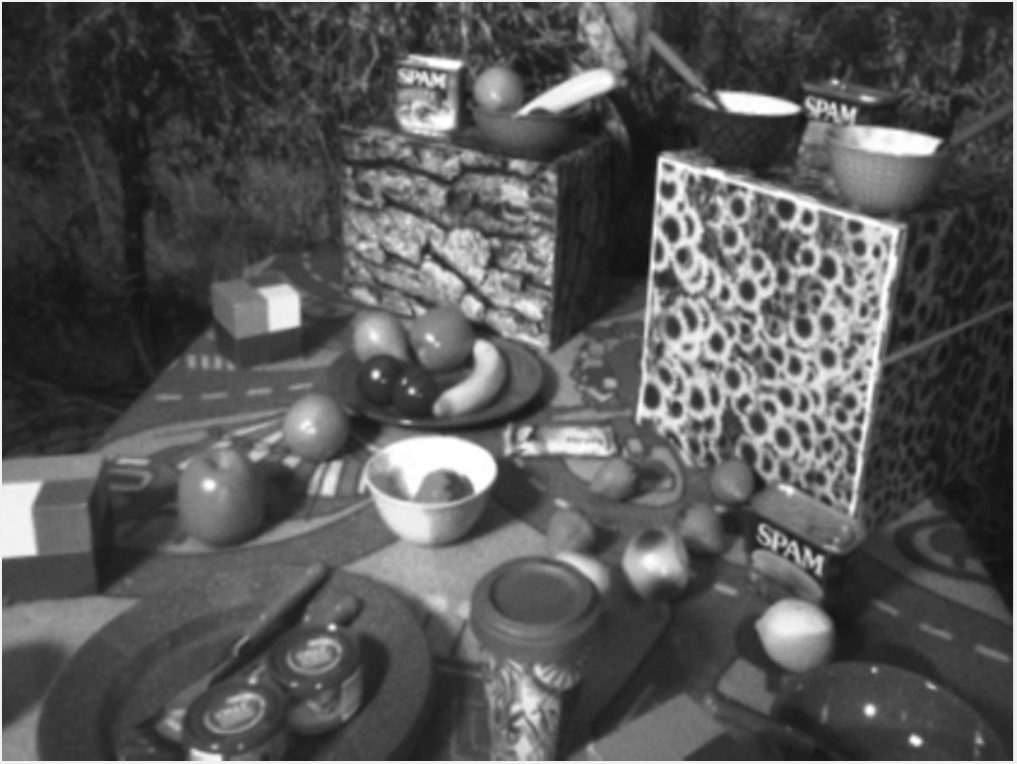}
		\hspace{-0.5 cm}
		& \includegraphics[width=0.21\textwidth]{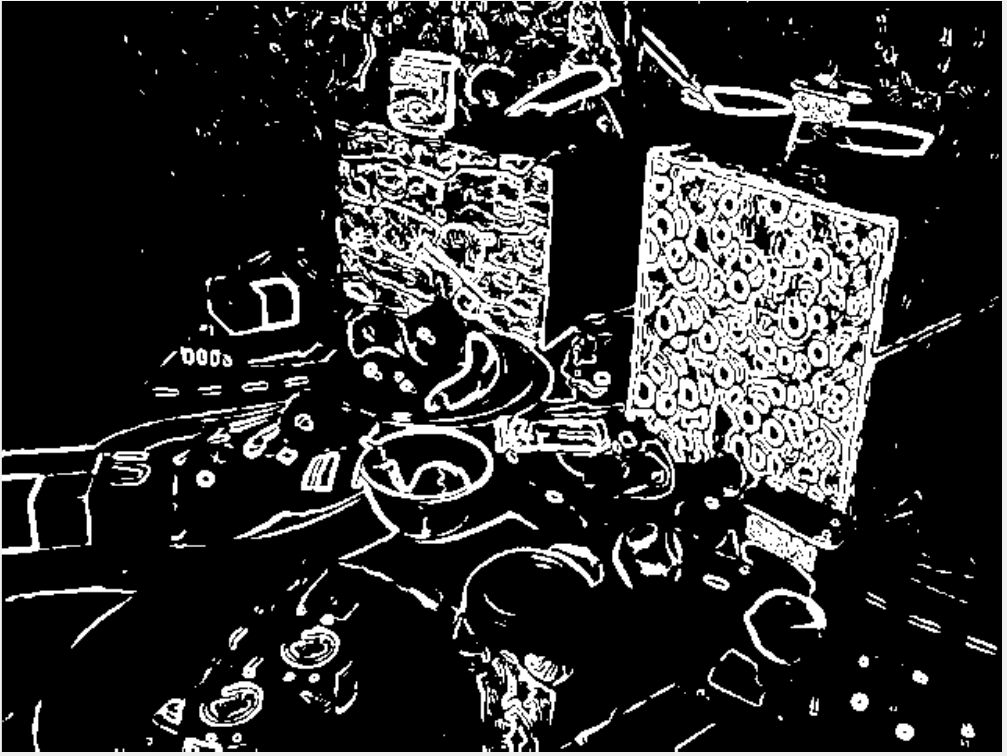}
		\hspace{-0.5 cm}
		& \includegraphics[width=0.21\textwidth]{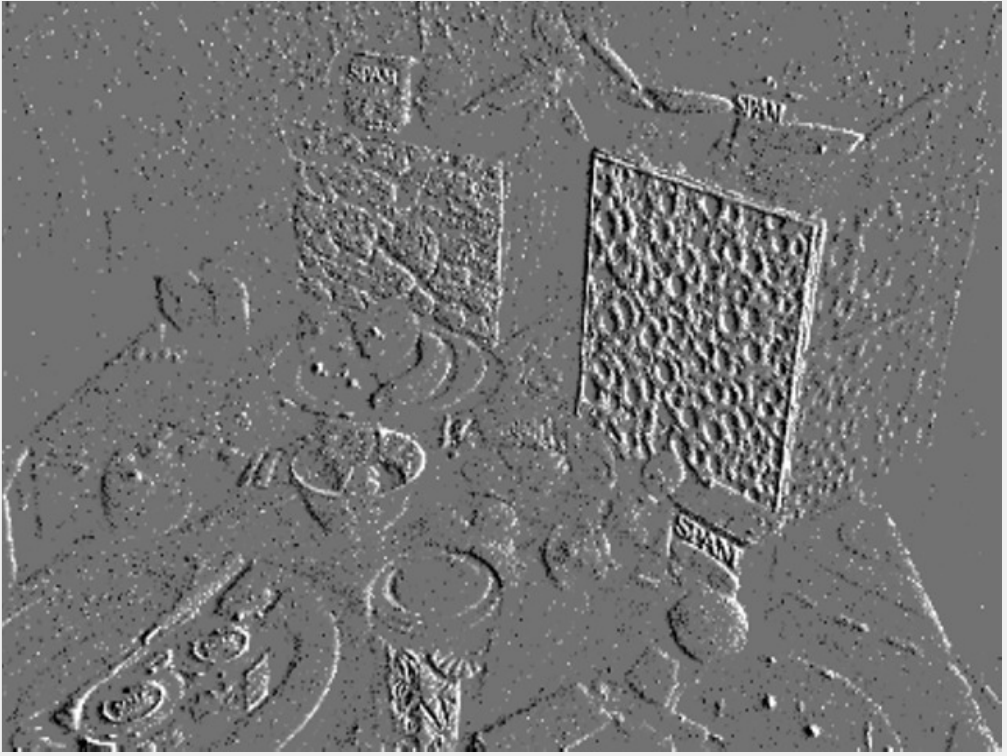}
		\hspace{-0.5 cm}
		& \includegraphics[width=0.21\textwidth]{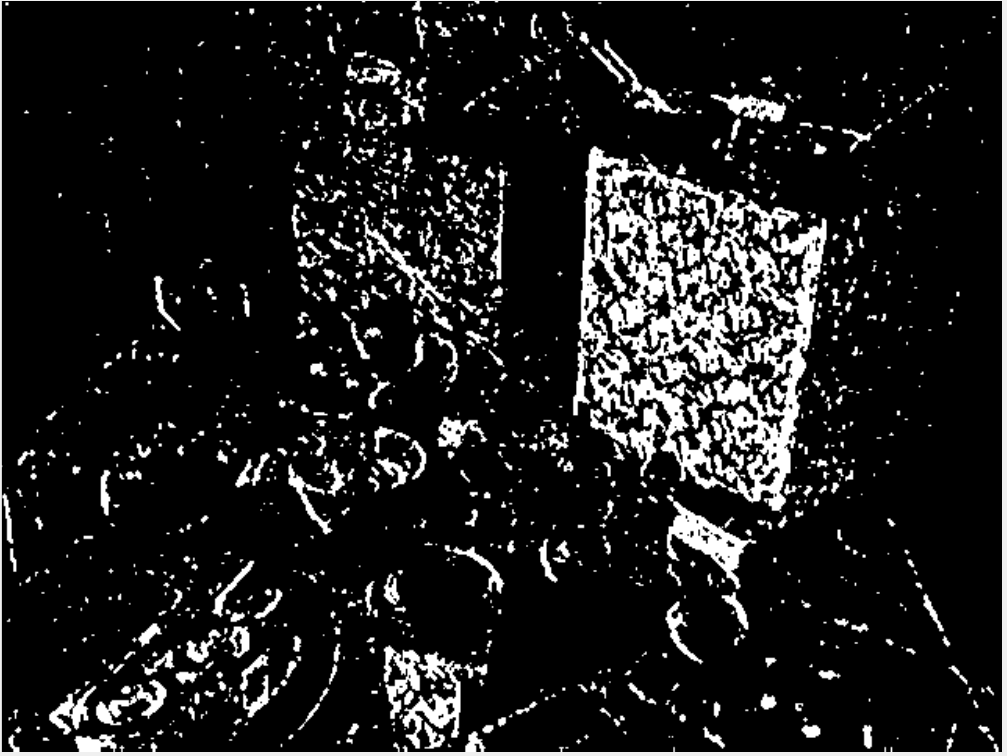} \\
		\hspace{-0.35 cm}
		(a) Image
		\hspace{-0.5 cm}
		& (b) $\mathbf{L}_b$
		\hspace{-0.5 cm}
		&(c) Event
		&(d) $\mathbf{E}_b$\\
	\end{tabular}
	\caption{Examples of binary edge maps.
		(a) Intensity image captured by the FLIR frame-based camera.
		The color has been converted to gray.
		(b) Binary edge map of (a).
		(c) Reconstructed event frame from event data captured by a Prophesee event camera.
		(d) Binary edge map of (c).}
	\label{fig:imeg}
\end{figure}

\noindent{\bf{Camera motions: }}
The camera motion is controlled by servoing UR5's end-effector to follow the pre-defined smooth trajectories to 0.1mm precision~\cite{UR5-specification}.
The camera trajectories in the dataset consist of, a circle, and three `8' shaped Lissajous trajectories. The three `8' shaped Lissajous trajectories include a planer trajectory parallel to the cameras' image plane, one with moderate variation along the cameras' optical axis, and a complex trajectory with significant variation along the cameras' optical axis.
In addition, a square trajectory of four pure translations is included.
Since using a robotic arm increases the reproducibility of the experiment, we are able to perform each trajectory six times at three different speeds, for two different lighting conditions and each of the three scenes described above.
The orientation of the stereo camera rig remains unchanged in all trajectories.  
Images used to reconstruct the ground truth 3D scene are captured from multiple views to reduce the effect of occlusion.

\subsection{Calibration}
\noindent{\bf{Stereo camera calibration:}}
We calibrate the intrinsic and extrinsic parameters of SHEF using the MATLAB stereo camera calibrator~\cite{stereoToolbox}. To continuously trigger events for calibration, we move a blinking checkerboard in front of the system and reconstruct intensity images using the event-based high pass filter~\cite{Scheerlinck18accv}.
We synchronize the two cameras by sending a trigger signal from the frame camera to the event camera.

\noindent{\bf{Hand-eye calibration:}} 
The SHEF rig is mounted on the end-effector of the UR5 manipulator. 
A hand-eye calibration procedure is performed to compute the rigid-body transformation from the frame camera to the end-effector so that the accurate camera poses with respect to the robot base can be directly computed from the manipulator's forward kinematics model.
During the calibration procedure, the robot end-effector, i.e the SHEF rig, is moved to observe a static flat calibration pattern from various viewpoints.

Denote the 6 DoF poses of the end-effector frame $\{H\}$, robot base frame $\{W\}$, and the calibration pattern $\{T\}$ as elements in the Special Euclidean Group $\mathrm{SE}(3)$. 
Then denote the pose of an arbitrary frame $\{B\}$ with respect to an arbitrary reference frame $\{A\}$ as $\idx{\bm{X}}{A}{}{B}{}$. The pose of the calibration pattern with respect to the robot base 
$\idx{\bm{X}}{W}{}{T}{} = \idx{\bm{X}}{W}{}{H}{}\idx{\bm{X}}{H}{}{C}{}\idx{\bm{X}}{C}{}{T}{}$. The same pose can be expressed from different viewpoints as
\begin{equation}
\idx{\bm{X}}{W}{}{H_i}{}\idx{\bm{X}}{H}{}{C}{}\idx{\bm{X}}{C_i}{}{T}{} = \idx{\bm{X}}{W}{}{H_j}{}\idx{\bm{X}}{H}{}{C}{}\idx{\bm{X}}{C_j}{}{T}{},
\label{eq:hand_eye_calib}
\end{equation}
where $i, j \in\{1, 2, 3 ....\}$ represent the index of different viewpoints. By rearranging Eq.\eqref{eq:hand_eye_calib}, we have
\begin{equation}
\bm{A}_{ij}\idx{\bm{X}}{H}{}{C}{} = \idx{\bm{X}}{H}{}{C}{}\bm{B}_{ij},
\label{eq:ax=ab}
\end{equation}
where $\bm{A}_{ij} = \idx{\bm{X}}{W}{}{H_j}{-1}\idx{\bm{X}}{W}{}{H_i}{},~ \bm{B}_{ij} = \idx{\bm{X}}{C_j}{}{T}{}\idx{\bm{X}}{C_i}{}{T}{-1}$. Both $\bm{A}_{ij}$ and $\bm{B}_{ij}$ are known; the former is computed from the forward kinematics model of the UR5 manipulator, and the latter is estimated from image frames. The transformation matrix from the camera to the robot base is estimated from Eq.\eqref{eq:ax=ab} using the hand-eye calibration method proposed in~\cite{liang2008hand_eye_calib}.
\begin{figure*}
	\vspace{1.5mm}
	\centering
	\begin{tabular}{cc|ccc}
		\includegraphics[width=0.183\textwidth,height=22.3mm]{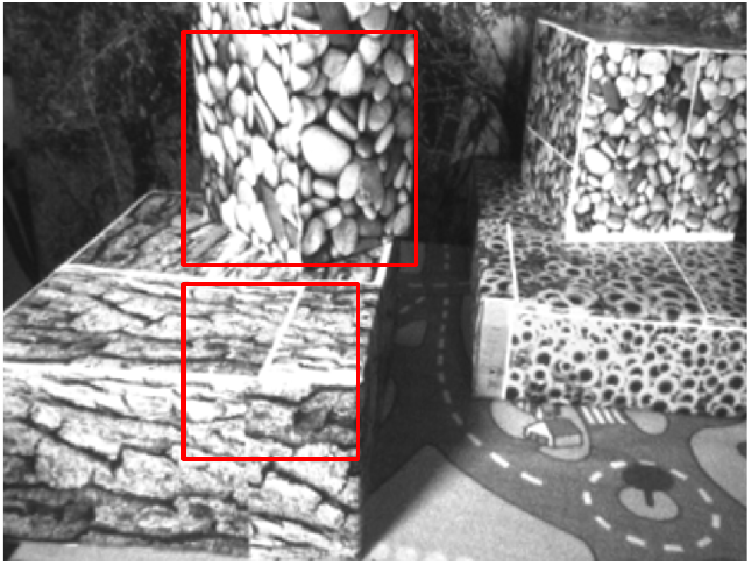} 
		& \includegraphics[width=0.183\textwidth]{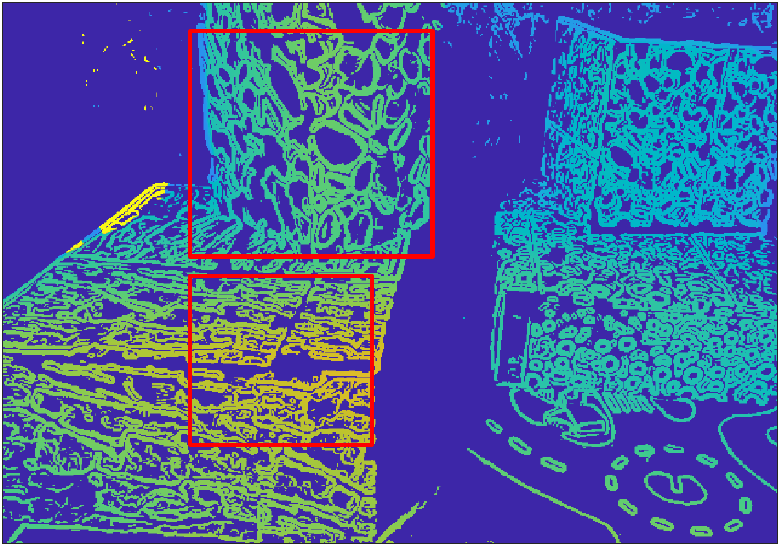}
		& \includegraphics[width=0.183\textwidth]{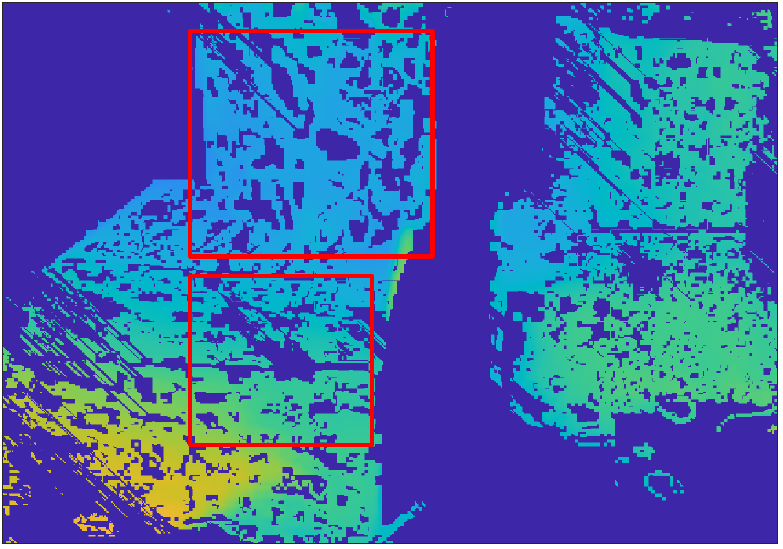}
		&\includegraphics[width=0.183\textwidth]{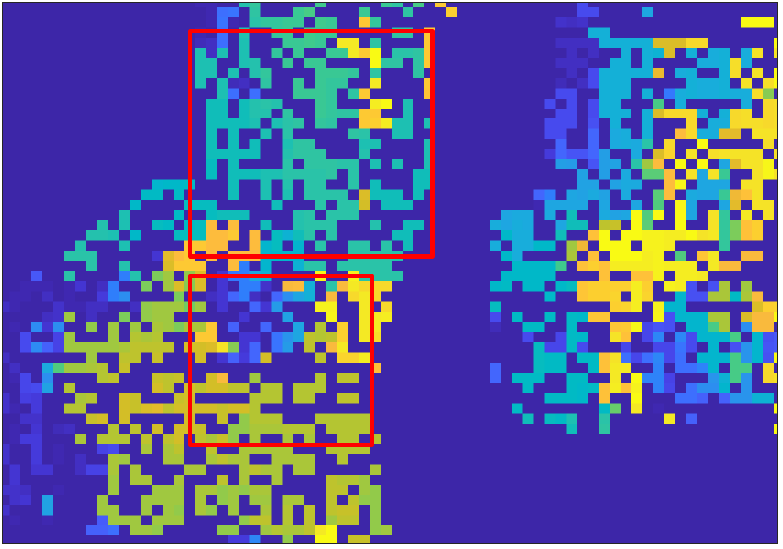} 
		&\includegraphics[width=0.183\textwidth]{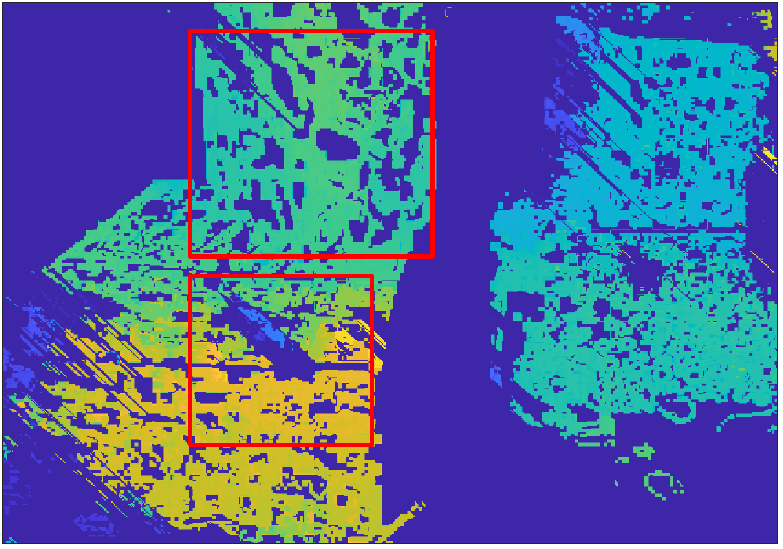}\\
		\includegraphics[width=0.183\textwidth,height=22.3mm]{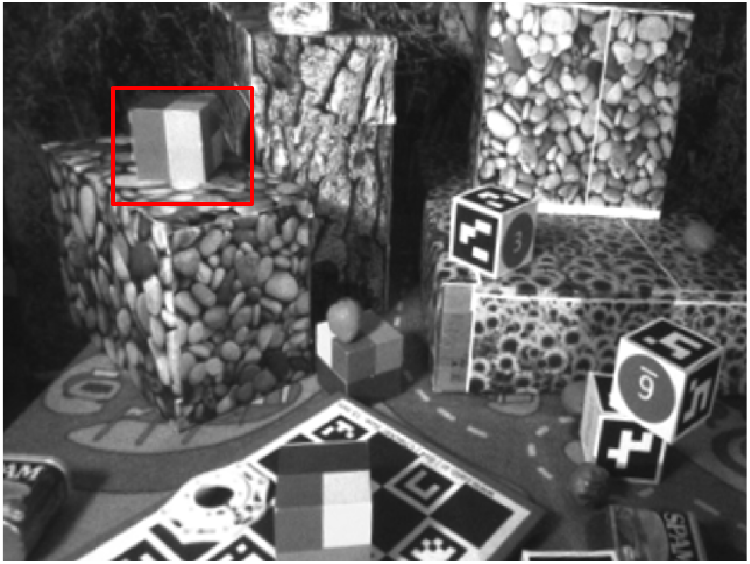} 
		& \includegraphics[width=0.183\textwidth]{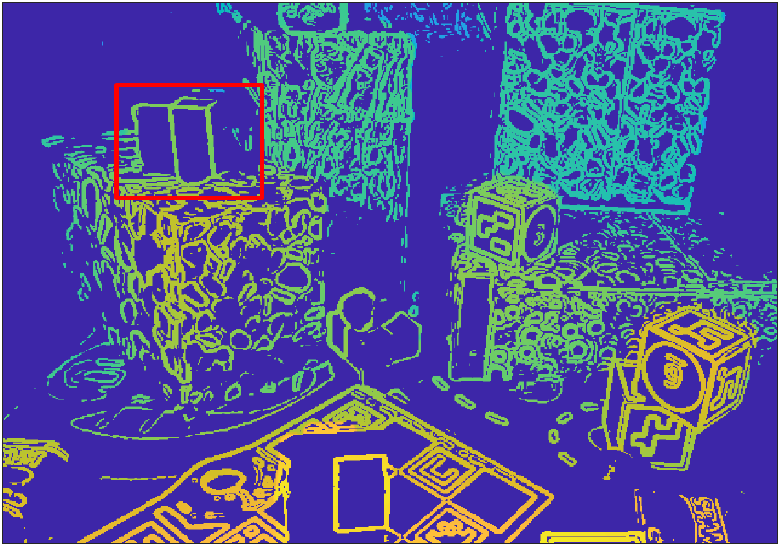}
		& \includegraphics[width=0.183\textwidth]{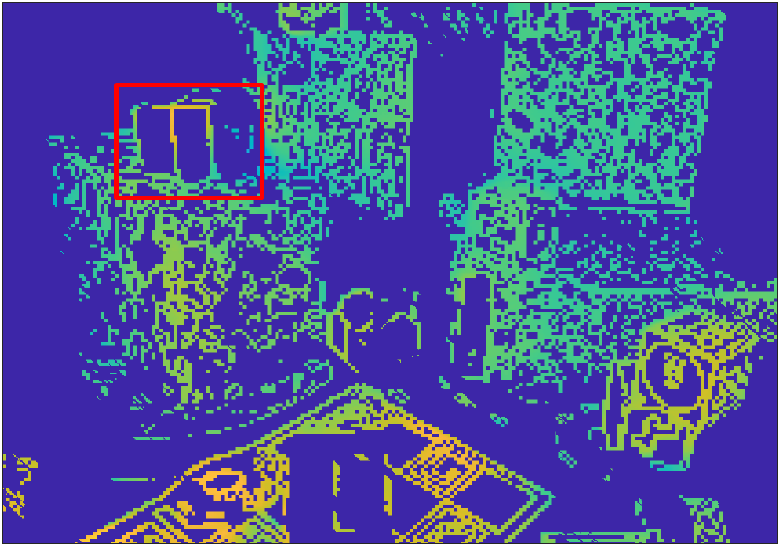}
		&\includegraphics[width=0.183\textwidth]{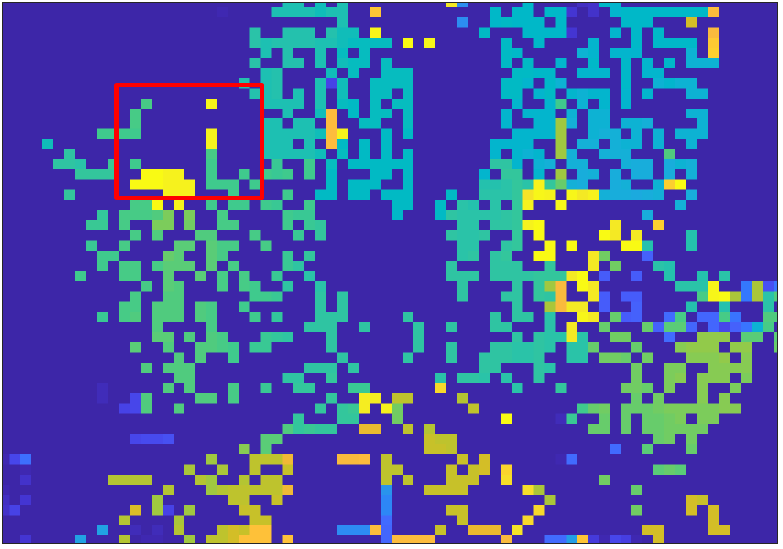} 
		&\includegraphics[width=0.183\textwidth]{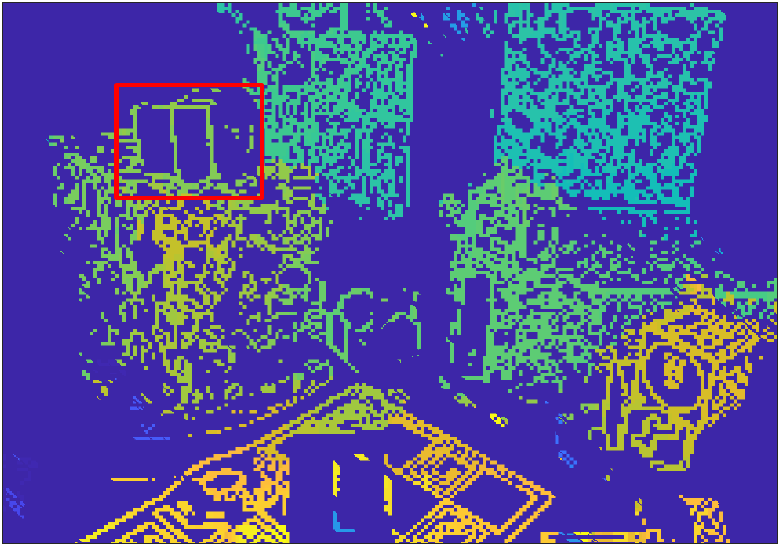}\\
		\includegraphics[width=0.183\textwidth,height=22.3mm]{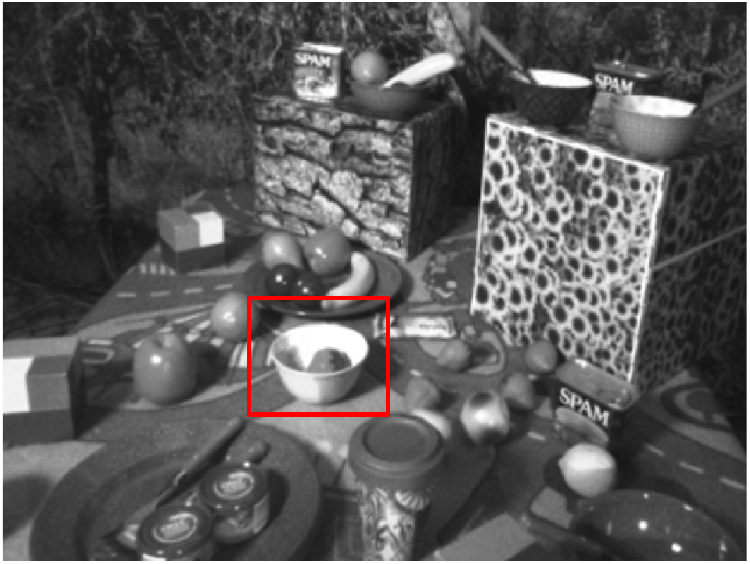} 
		& \includegraphics[width=0.183\textwidth]{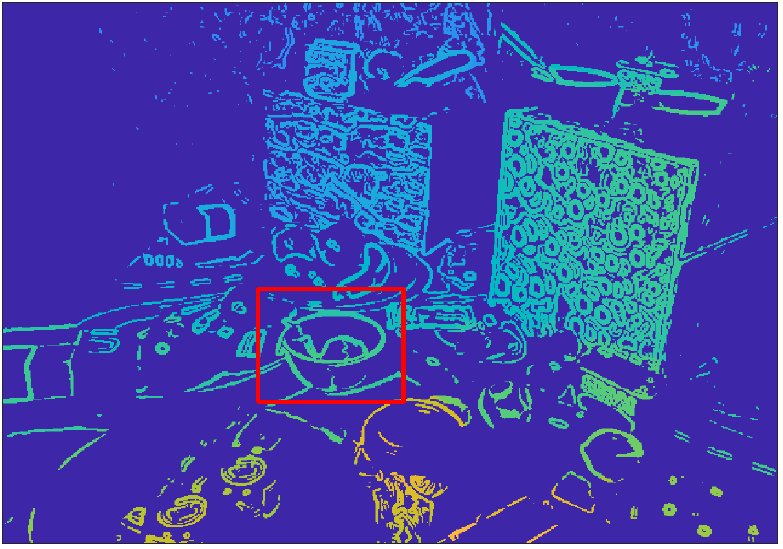}
		& \includegraphics[width=0.183\textwidth]{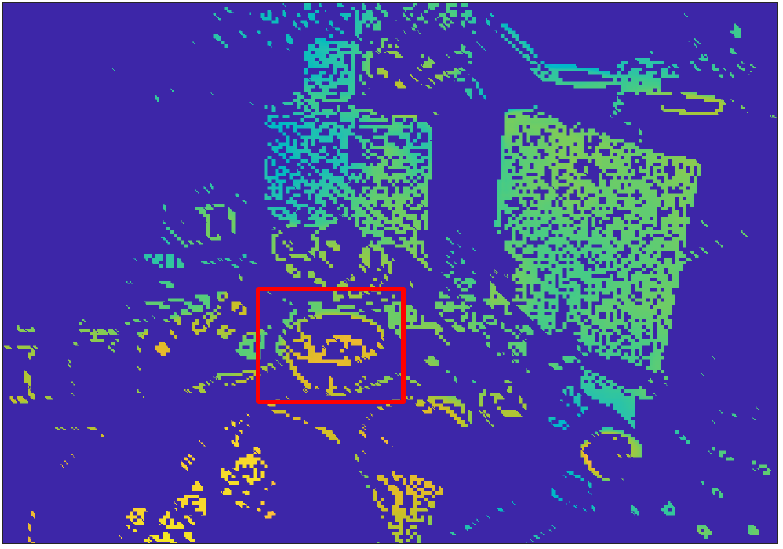}
		&\includegraphics[width=0.183\textwidth]{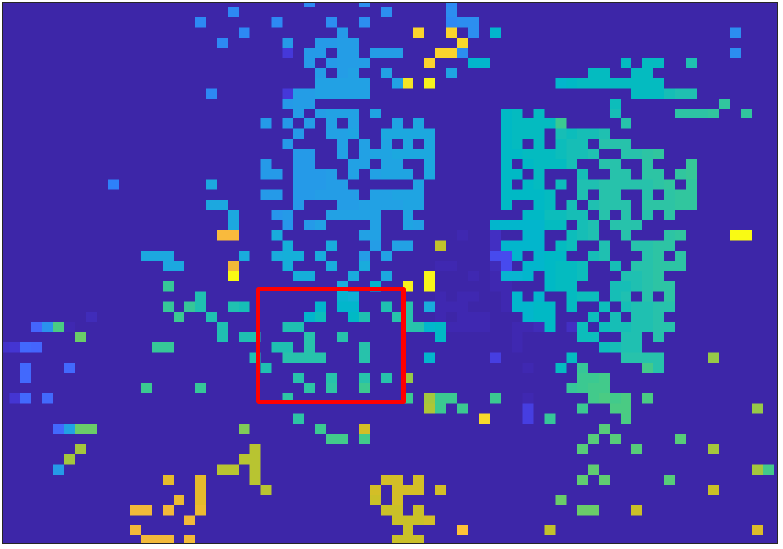} 
		&\includegraphics[width=0.183\textwidth]{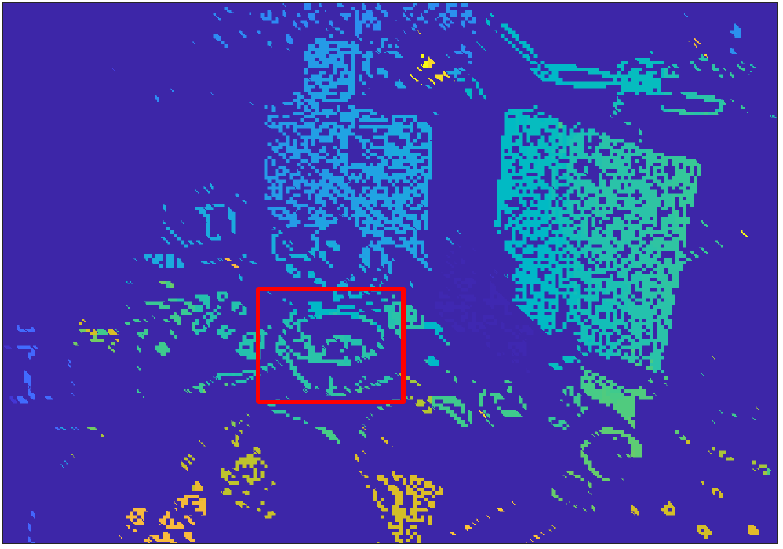}\\
		(a) Image
		\hspace{-0.349 cm}
		&(b) Edges $\mathbf{D}_{GT}$ 
		\hspace{-0.349 cm}
		&(c) $\mathbf{D}_{AA}$ \cite{xu2020aanet}
		\hspace{-0.349 cm}
		&(d) Ours $\mathbf{D}_s$
		\hspace{-0.349 cm}
		&(e) Ours $\mathbf{D}_p$
	\end{tabular}
	\caption{Examples of the comparison of our results in three setups. (a) Intensity image. (b) Ground truth disparity map at intensity image edges.
		(c) Dense disparity map $\mathbf{D}_{AA}$ computed by the state-of-the-art stereo-frame-based disparity method AANet, using the input of intensity frames and pure-event reconstruction by the state-of-the-art algorithm ECNN~\cite{Stoffregen20eccv}. 
		(d) Our sparse disparity map $\mathbf{D}_s$.
		(e) Our sparse disparity map $\mathbf{D}_p$ with the coarse-to-fine framework.
		Disparity maps are color coded, from blue (far) to yellow (close), in the range of 0-80 pixels. Area of significant differences is highlighted with red rectangle.
	}
	\label{fig:all}
\end{figure*}

\noindent{\bf{Ground truth: }} The dataset requires the ground truth depth map associated with each frame.
For each dataset environment, we use the FLIR camera to take around 100 RGB images from different views and compute camera poses from the manipulator. Then we build a dense point cloud and a 3D reconstruction model for each set up (see Fig.~\ref{fig:gt}) using a commercial software \textit{3DFLOW Zephyr}~\cite{3dflow}.
With the known camera pose and intrinsic parameter for each frame in our dataset, we compute the camera ray for each pixel in the frame and find the corresponding 3D point along the ray. 
Then, we transform the 3D point cloud into a 2D ground truth depth map for each frame.

\section{STEREO DISPARITY ESTIMATION}
Different from the standard stereo frame-based disparity estimation problem, our stereo hybrid event-frame system provides an event stream at one view and an intensity frame at the other view. 
As the sensing modality of the event data is different from its corresponding intensity frame data, correspondence-based algorithms that rely on matching cost functions cannot be applied directly.
In this section, we provide a stereo disparity estimation baseline algorithm that 
correlate the edge information extracted from event data and frames. 
A coarse-to-fine framework is also applied 
to compute denser and smoother disparity.

\noindent{\bf{Binary edge images: }}
To estimate the disparity of event and frame image at a certain frame timestamp $t$,
we first use high pass filter~\cite{Scheerlinck18accv} to reconstruct intensity images from events at time $t$ with a relatively high cut-off frequency (120 rad/s), resulting in a short filter memory and only preserving high frequency edge information in a deterministic and asynchronous manner. 
On the reconstructed image, we perform non-maximal suppression to generate a binary edge map $\mathbf{E}_b$. 
The binary edge maps are generated at the same timestamp as the paired frame camera. 
For the frame data, we compute an edge map using classical $3 \times 3$ Sobel filters and then apply non-maximal suppression to generate the corresponding binary edge map $\mathbf{L}_b$  (Fig.~\ref{fig:imeg}).

\noindent{\bf{Cross-correlation: }}
We slide a window along the epipolar line in the right-hand binary edge image and calculate the cross-correlation between the reference window in the left-hand binary edge image. 
By minimizing the matching cost, we get a sparse disparity map $\mathbf{D}_s\in
{\rm I} \! {\rm R}^{H\times W}$. Here, $H$ and $W$ are the image height and width.

We estimate the disparity of a pixel $\mathbf{x}=(x,y)$ by solving,
\begin{equation}
\begin{aligned}
\mathbf{D}_s(\mathbf{x}) = \min_{\triangle x}\lambda\{e(\mathbf{x}+\triangle x)\}\ .  
\end{aligned}
\end{equation}
The matching cost $e(\cdot)$ is given by,
\begin{equation}
\begin{aligned}
e(\mathbf{x}&+\triangle x)\\
&=\sum_{i,j\in\mathcal{W}}\mathbf{L}_b(x+i,y+j) \odot \mathbf{E}_b(x+i+\triangle x,y+j)\ ,  
\end{aligned}
\end{equation}
where $\mathbf{D}_s$ denotes the sparse disparity map, $(i,j)$ are the pixels in the sliding window $\mathcal{W}$, and $\lambda=-1$ is the weight parameter. Here, $\triangle x$ is the estimated disparity.

\noindent{\bf{Coarse-to-fine framework: }}
To make our sparse disparity map denser and smoother, we implement a traditional coarse-to-fine framework using an image pyramid with 5 levels and a scale factor of 0.9. 
Disparity estimation then operates coarse-to-fine over the defined pyramids by 
\begin{equation}
\begin{aligned}
\mathbf{D}_p^{k}= \rm{I_1} \mathbf{D}_p^{k+1}\uparrow + \rm I_2 \min_{\triangle x}\lambda\{e(\mathbf{x}+\triangle x)\}_k \ ,  
\end{aligned}
\end{equation}
where $k \in \{1,\cdots,5\}$ is the pyramid level, $\uparrow$ denotes upsampling by the scale factor, and $\rm I_1$ and $\rm I_2$ are weight matrices. We represent the improve disparity map as $\mathbf{D}_p$. Note that, \resizebox{0.093\textwidth}{!}{ $\rm I_* \in {\rm I} \! {\rm R}^{H\times W} $}, where $\rm I_1 + \rm I_2 = \bf 1$ and all elements in the two matrices are in the set $\{0,0.5,1\}$. 
Finally, we remove disparity outliers that do not have sufficient surrounding support pixels, where the local estimated disparity should be similar. 

\section{Experiment}

\subsection{Experimental Setup and Main Results. }
We evaluate our disparity estimation algorithm on three scenes: \texttt{simple boxes}, \texttt{complex boxes} and \texttt{picnic}.
For each scene, we evaluate on 50 event-frame image pairs under the same motion speed (fast), lighting condition (bright) and trajectory (circle).
The algorithm is run offline in batch mode. 
To evaluate the quality of the disparity map, we first compute the ground truth disparity $\mathbf{D} = \frac{\mathbf{b}\mathbf{f}}{\mathbf{d}}$, where $\mathbf{b}$, $\mathbf{f}$, $\mathbf{d}$ denotes
the camera baseline, focal length and depth respectively.

\noindent{\bf Comparison: }
Since there is no similar existing dataset and algorithm, to prove the effectiveness of our disparity estimation algorithm, we compare our method with the state-of-the-art stereo frame-based matching Adaptive Aggregation Network (AANet)~\cite{xu2020aanet} using our dataset. 
We first use the state-of-the-art pure-event reconstruction method ECNN~\cite{Stoffregen20eccv} to generate an event image to each 
corresponding intensity image, and then estimate dense disparity from the stereo image pairs using the AANet.
We represent the dense disparity map computed by AANet as $\mathbf{D}_{AA}$.

The quality of sparse $\mathbf{D}_s$ using cross-correlation and the improved $\mathbf{D}_p$ with an image pyramid are separately evaluated against the ground-truth; only the pixels with estimated disparity are considered.
We evaluate the disparity of $\mathbf{D}_{AA}$
on the edge pixels of $\mathbf{D}_p$ in Table \ref{tab:DPD-disp} and compare the dense disparity $\mathbf{D}_{AA}$ with our completed dense disparity $\mathbf{D}_c$ (discussed in \S \ref{sec:Disparity completion}) in Table \ref{tab:DPD-disp-dense}. 
In the quantitative evaluation of all methods, we ignore pixels on the right and left image boundaries because they do not have corresponding matching pixels in the other stereo image pair.

\noindent{\bf Evaluation metrics: }
We evaluate the disparity estimation quality using three metrics: RMSE (Root-Mean-Square Error), bad-p and the inlier ratios~\cite{monodepth17}. 
`bad-p' is the standard metric of KITTI~\cite{menze2015object} which computes the percentage of bad pixels averaged over pixels with the estimated disparity of all testing images.
For our experiment, we consider a pixel to be correctly estimated if the disparity end-point error is $<5$ pixels or $<5\%$. 
The $\delta$ inlier ratios measure the percentage of the disparity error within the preset maximal mean relative error of $\delta_i = 1.25^i$, where $i\in \{1,2,3\}$.

\noindent{\bf Main results: }
In Figure \ref{fig:all}, we evaluate
our sparse disparity map $\mathbf{D}_s$, the improved disparity map $\mathbf{D}_p$ (using a coarse-to-fine framework) and $\mathbf{D}_{AA}$ at edge pixels of $\mathbf{D}_p$.
Because the binary edges from the event camera are very noisy and they are highly dependent on camera motion (see Fig. \ref{fig:imeg}), some of the edges cannot be captured by $\mathbf{D}_s$.
Using the coarse-to-fine framework, the disparity of each pixel in $\mathbf{D}_p$ contains a local average corresponding to its neighborhood pixels that average out the noise and takes advantage of the spatial information.
Figure \ref{fig:all} shows that our coarse-to-fine framework leads to a smoother and more accurate disparity map. 
Especially in the \texttt{simple boxes} dataset (the first row of Fig. \ref{fig:all}), the image pyramid helps $\mathbf{D}_p$ to obtain more accurate disparity estimation (yellow to green).

Table \ref{tab:DPD-disp} shows that our coarse-to-fine result $\mathbf{D}_p$ achieves more than 51\% relative improvement in RMSE and 55\% relative improvement in bad-p respectively versus $\mathbf{D}_s$.
It also outperforms the learning-based stereo method AANet~\cite{xu2020aanet} (evaluated on edges) with a significant margin in all three metrics.
Since event cameras only report intensity changes at textured areas, ECNN \cite{Stoffregen20eccv} relies on regularisation and spatial priors to reconstruct a smooth and dense image.
This will distort the edge information in the reconstructed image and degrade the performance of AANet.
For example, in Fig.~\ref{fig:all} highlighted by the red bounding box, 
$\mathbf{D}_{AA}$ failed to estimate disparity correctly but our $\mathbf{D}_p$ achieves the more precise disparity estimation.

\begin{table}[]
	\centering
	\caption{ \label{tab:DPD-disp}
		\em Quantitative analysis of disparity estimation on our SHEF dataset evaluated on edge pixels. 
	}
	\vspace{3mm}
	\aboverulesep=0ex
	\belowrulesep=0ex
	\begin{adjustbox}{width=0.45\textwidth}
		\begin{tabular}{lccccc}
			\midrule
			\midrule
			& RMSE $\downarrow$ & bad-p $\downarrow$ & $\delta_1$ $\uparrow$ & $\delta_2 $ $\uparrow$
			& $\delta_3$ $\uparrow$ \\
			\hline
			$\mathbf{D}_{AA}$ (edge pixels) & 11.07 & 52.72\%   & 0.7262   & 0.8980    & 0.9193   \\ \hline
			Ours $\mathbf{D}_s$ & 17.24    & 40.08\%    & 0.6967    & 0.7641   & 0.8530  \\ \hline   
			Ours $\mathbf{D}_p$ & \textbf{8.38} & {\bf 17.76\%}   &  \textbf{0.9157}   & \textbf{0.9334}   & \textbf{0.9527}
			\\ \midrule
			\midrule
		\end{tabular}
	\end{adjustbox}
\end{table}

\begin{figure}[t]
	\vspace{5mm}
	\centering
	\includegraphics[width=.3\columnwidth,trim=30 0 80 0, clip]{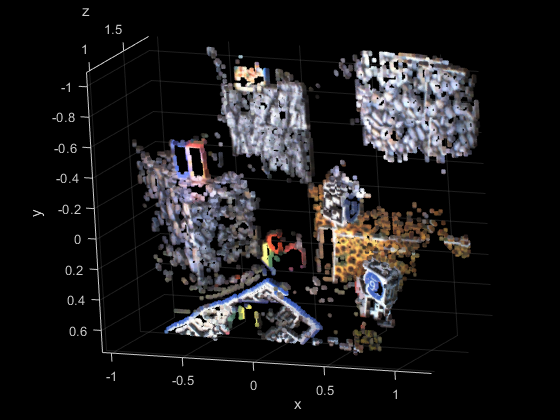}
	\caption{Reconstructed 3D point cloud of a part (visible image) of scene b) computed from estimated disparity $\mathbf{D}_p$.}
	\label{fig:reconstructed_pointcloud}
\end{figure}

To provide a visualisation of the results, we plot the 3D point cloud from the computed disparity ($\mathbf{D}_p$) in Fig.~\ref{fig:reconstructed_pointcloud}, with point colour taken from the frame image. 
The resulting 3D scene is a reasonable geometric representation of a part of the scene in Fig.~\ref{fig:gt}(b) and it would be sufficient to undertake robotic tasks like grasping. 

\subsection{Disparity completion.}\label{sec:Disparity completion}
Compared to a stereo camera system with two event sensors, our SHEF camera system also provides RGB image intensities.  
To demonstrate the potential of our system and for better visualization, we use a sparse-to-dense disparity completion Network (DCNet) to combine the sparse disparity data and dense images, using both the event and intensity cues.
\begin{figure}
	\vspace{2mm}
	\centering
	\begin{tabular}{ccc}
		\hspace{-0.35 cm}
		\includegraphics[width=0.21\textwidth]{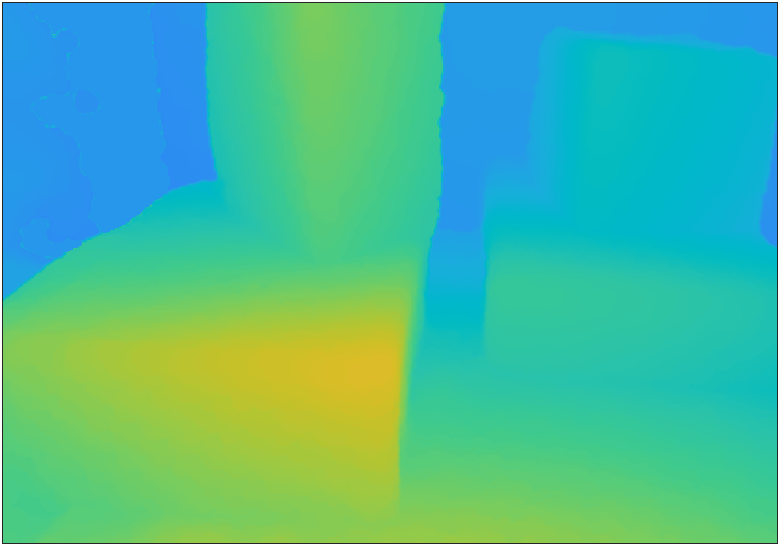} 
		\hspace{-0.35 cm}
		& \includegraphics[width=0.21\textwidth]{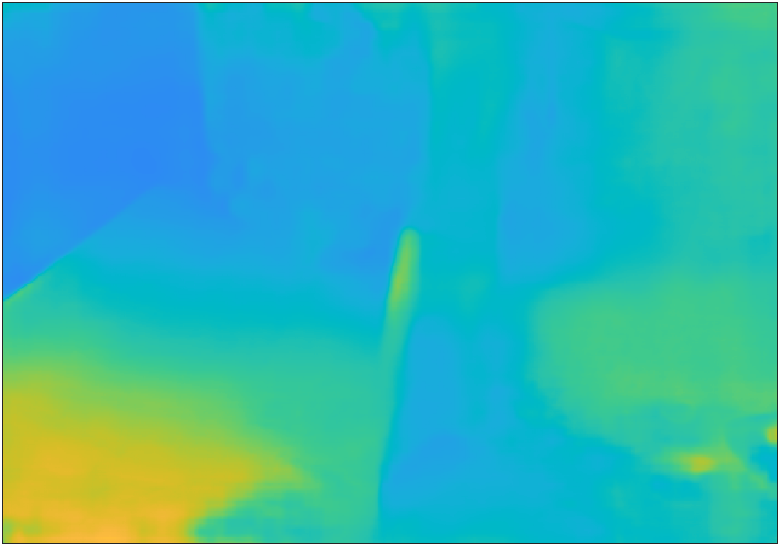} 
		\hspace{-0.35 cm}
		& \includegraphics[width=0.21\textwidth]{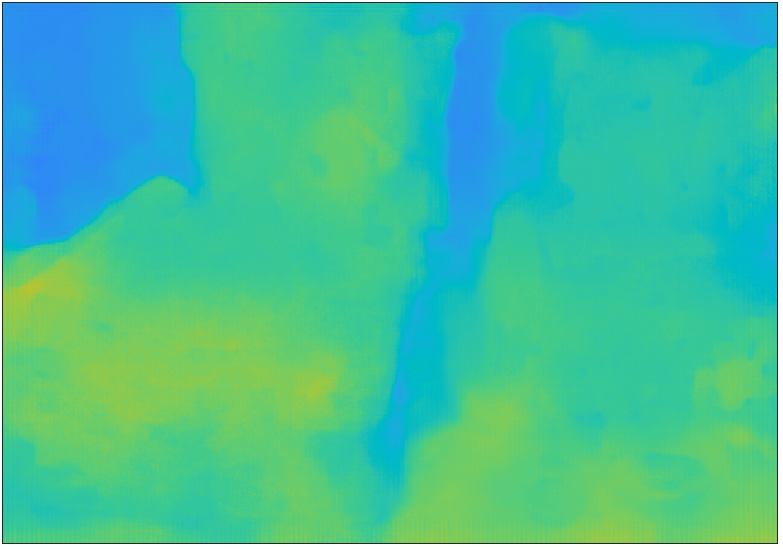} \\
		\hspace{-0.35 cm}
		\includegraphics[width=0.21\textwidth]{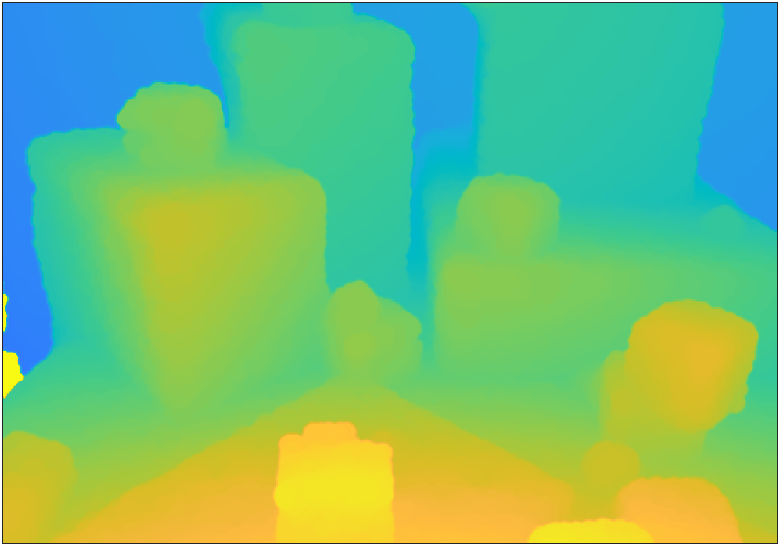} 
		\hspace{-0.35 cm}
		& \includegraphics[width=0.21\textwidth]{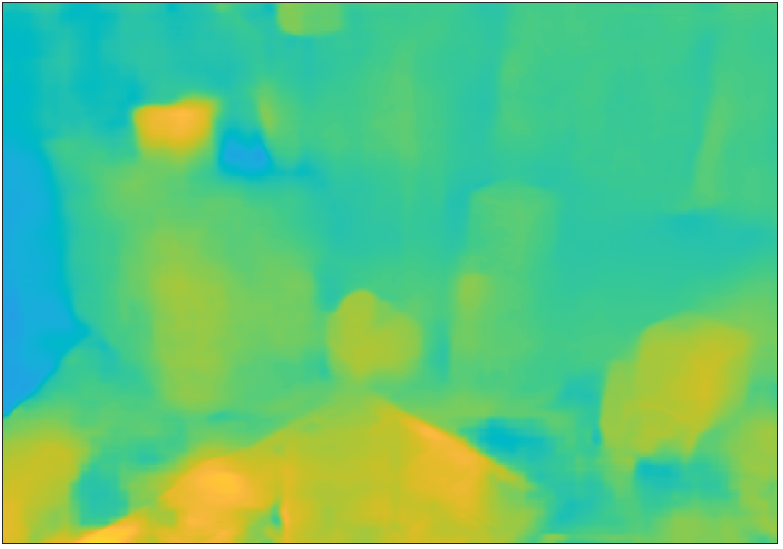} 
		\hspace{-0.35 cm}
		& \includegraphics[width=0.21\textwidth]{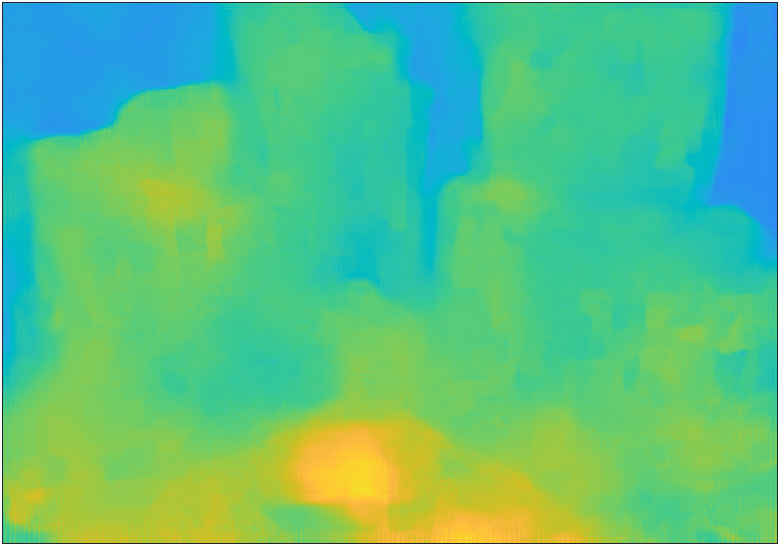}
		\\
		\hspace{-0.35 cm}
		\includegraphics[width=0.21\textwidth]{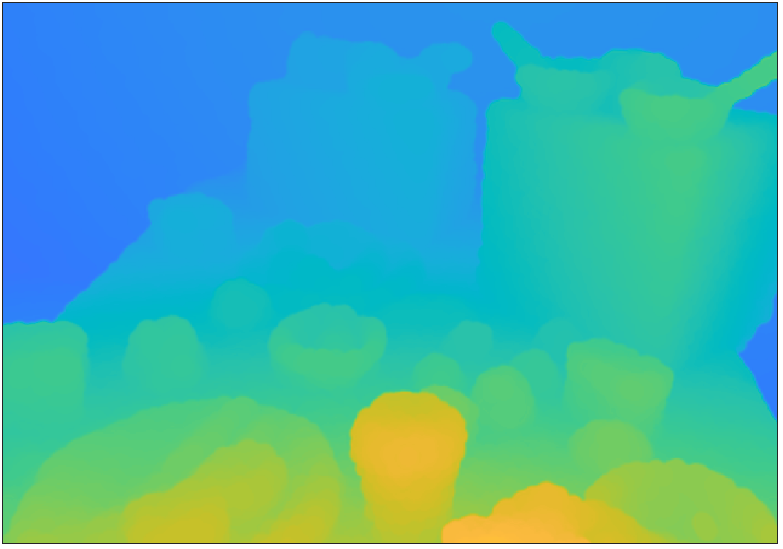} 
		\hspace{-0.35 cm}
		& \includegraphics[width=0.21\textwidth]{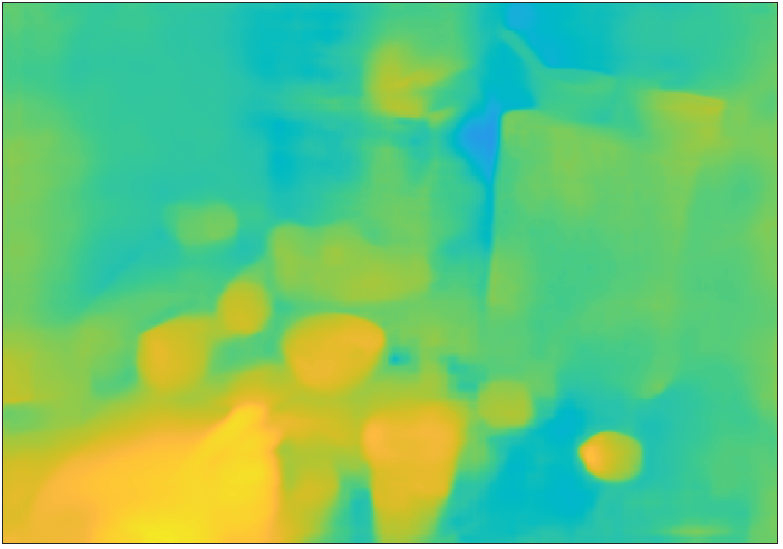} 
		\hspace{-0.35 cm}
		& \includegraphics[width=0.21\textwidth]{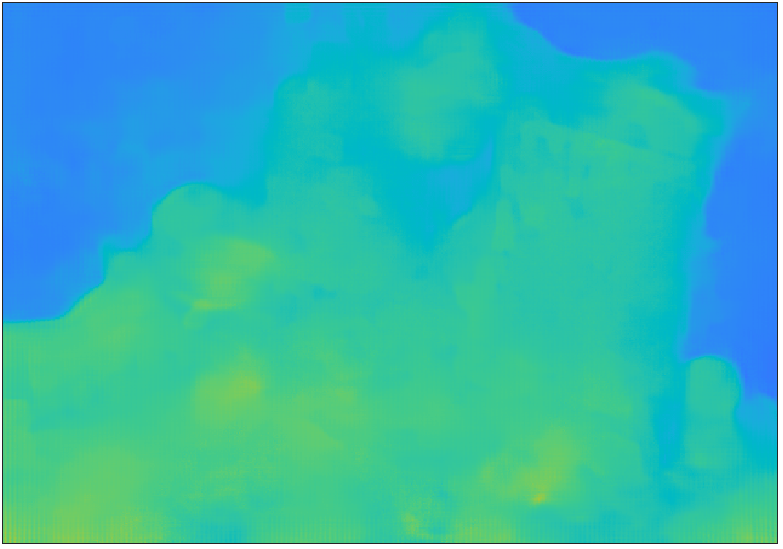} \\
		\hspace{-0.35 cm}
		(a) $\mathbf{D}_{GT}$
		\hspace{-0.35 cm}
		& (b) $\mathbf{D}_{AA}$ \cite{xu2020aanet} 
		\hspace{-0.35 cm}
		&(c) Ours $\mathbf{D}_c$   \\
	\end{tabular}
	\caption{Examples of the disparity completion results. (a) The ground truth dense disparity map. (b) Disparity map $\mathbf{D}_{AA}$ computed by AANet. The inputs of the AANet are the reconstructed images from the event stream and the intensity images. (c) Ours dense disparity computed by DCNet. 
		Note that AANet and DCNet have been fine-tuned or trained on our generated training data. 
		Disparity maps are color codes, from blue (far) to yellow (close), in the range of 0-80 pixels.}
	\label{fig:dense}
\end{figure}

\noindent{\bf{Network architecture: }}
Our DCNet $g(\cdot)$ with parameters $\mathcal{G}$ first extract feature maps from image and event, and calculate a 3D cost volume. Then, we use the method proposed by \cite{saikia2019autodispnet,pan2021dual} to regress the disparity maps. 

The input of our DCNet is the event frame $\mathbf{E}_b$, the image $\mathbf{L}$, and the pre-estimated sparse disparity map $\mathbf{D}_p$. The output of our DCNet is the estimated disparity map ${\mathbf{D}_c} = g(\mathbf{L},\mathbf{E}_b,\mathbf{D}_p;\mathcal{G})$. We use ground-truth disparity map $\mathbf{D}$ for training, and adopt the widely used smooth $\ell_1$ loss $\mathcal{S}(\cdot)$~\cite{girshick2015fast} to penalizes the differences between $\mathbf{D}$ and $\mathbf{D}_c$. Our loss function $\mathcal{L}$ is defined as
\begin{equation} \label{eq:loss2}
\begin{aligned}
{\mathcal{L}} = \frac{1}{N}\sum_{x,y} \mathcal{S}(\mathbf{D}(x,y)-{\mathbf{D}_c}(x,y)) \ , \\
\end{aligned}
\end{equation}
where $N$ is the number of pixels.

The training data is based on the Middlebury dataset~\cite{scharstein2014high}, which provides stereo image pairs, camera parameters, and the associated ground-truth disparity map. We take the left-view as the reference, and use the right-view image to generate the event data. With a given rotation and translation matrix, we reproject the right-view image to a new view (horizontal shifting only), and then generate the event stream. Our training data only includes 450 event-frame pairs. For a fair comparison, both AANet and DCNet is trained and fine-tuned on the same dataset.

\noindent{\bf{Implementation details: }} Our network is implemented in Pytorch and is trained from scratch using the Adam
optimizer~\cite{kingma2014adam} with a learning rate of $10^{-4}$ and a batch
size of $10$. Our model is trained on a single NVIDIA Titan XP GPU. 

\noindent{\bf{Result: }}
We demonstrate the dense disparity evaluation of our result $\mathbf{D}_c$ by DCNet and $\mathbf{D}_{AA}$ by AANet in Fig.~\ref{fig:dense}.
From both dense disparity result, $\mathbf{D}_{AA}$ and our $\mathbf{D}_c$, the outlines of the objects are distinguishable, but AANet leads to large areas with wrong disparity estimation. For example, in the \texttt{simple boxes} scene (first line in Fig.~\ref{fig:dense}), $\mathbf{D}_{AA}$ has a smooth dense disparity of the left box, but wrongly estimate the disparity of the entire object.
This explains the higher errors in $\mathbf{D}_{AA}$ for all three metrics.

The Table~\ref{tab:DPD-disp-dense} shows that the large majority of the pixels of $\mathbf{D}_{AA}$ and our $\mathbf{D}_c$ has the maximal mean relative error of $\delta_2$, indicating only a small amount of outliers in the disparity map are introduced from both the methods.
However, $\mathbf{D}_{AA}$ leads to only around $64\%$ pixels with the maximal mean relative error of $\delta_1$ and more than $56\%$ pixels have disparity error larger than 5 pixels or 5$\%$,
which may be caused by the fact that $\mathbf{D}_{AA}$ has more pixels with imprecise disparity around the texture-less region.
Comparing to $\mathbf{D}_{AA}$, our $\mathbf{D}_c$ is more reliable and it has around $16\%$ smaller RMSE.

\section{CONCLUSIONS}
In this paper, we discuss the advantages of the stereo hybrid event-frame (SHEF) camera configuration and introduce a high-quality dataset targeted at stereo hybrid event-frame disparity estimation in near field scenes.
We provide a baseline disparity estimation method that computes cross-correlation between binary edge image pairs of event data and frames with a coarse-to-fine framework.
Our method outperforms the state-of-the-art stereo matching method AANet on the proposed dataset.

\begin{table}[]
	\centering
	\caption{ \label{tab:DPD-disp-dense}
		Quantitative analysis of dense disparity estimation on our SHEF dataset. 
	}
	\vspace{3mm}
	\aboverulesep=0ex
	\belowrulesep=0ex
	\begin{adjustbox}{width=0.47\textwidth}
		\begin{tabular}{lccccc}
			\midrule
			\midrule
			& RMSE $\downarrow$ & bad-p $\downarrow$ & $\delta_1$ $\uparrow$ & $\delta_2 $ $\uparrow$
			& $\delta_3$ $\uparrow$ \\
			\hline
			$\mathbf{D}_{AA}$~\cite{xu2020aanet} & 10.66 & 56.61\%   & 0.6424   & 0.8927     & 0.9563   \\ \hline
			Ours $\mathbf{D}_c$ &  \textbf{8.99} & \textbf{31.71\%}   & \textbf{0.8598}   &{\bf 0.9564}    &\textbf{0.9791} \\
			\midrule
			\midrule
		\end{tabular}
	\end{adjustbox}
\end{table}

\section{ACKNOWLEDGMENTS}
The authors would like to thank Pieter van Goor for helping ground truth depth collection, and thank Prophesee for providing the event camera that was used in the work.

\clearpage
\bibliographystyle{ieee_fullname}
\bibliography{template_arxiv}

\end{document}